\documentclass{article}

     \PassOptionsToPackage{numbers, sort, compress}{natbib}
      \usepackage[preprint]{neurips_2020}

\usepackage[utf8]{inputenc} % allow utf-8 input
\usepackage[T1]{fontenc}    % use 8-bit T1 fonts
\usepackage{hyperref}       % hyperlinks
\usepackage{url}            % simple URL typesetting
\usepackage{booktabs}       % professional-quality tables
\usepackage{amsfonts}       % blackboard math symbols
\usepackage{nicefrac}       % compact symbols for 1/2, etc.
\usepackage{microtype}      % microtypography

\usepackage{graphicx}
\usepackage{subfigure}
\usepackage{amsmath}
\usepackage{amssymb}
\usepackage{multirow}
\usepackage{algorithm}
\usepackage{algorithmic}
\usepackage{color}
\usepackage{makecell}

\title{Learning to Match Distributions for Domain Adaptation}

\author{%
	Chaohui Yu$^{1}$, Jindong Wang$^{2}$\thanks{The first two authors contributed equally. Correspondence to: jindong.wang@microsoft.com}, Chang Liu$^2$, Tao Qin$^2$,\\ \textbf{Renjun Xu$^3$, Wenjie Feng$^1$, Yiqiang Chen$^1$, Tie-Yan Liu$^2$} \\
	$^1$ICT, CAS $^2$Microsoft Research $^3$Zhejiang University
}
\begin{document}

\maketitle

\begin{abstract}
When the training and test data are from different distributions, domain adaptation is needed to reduce dataset bias to improve the model's generalization ability. Since it is difficult to directly match the cross-domain joint distributions, existing methods tend to reduce the marginal or conditional distribution divergence using predefined distances such as MMD and adversarial-based discrepancies. However, it remains challenging to determine which method is suitable for a given application since they are built with certain priors or bias. Thus they may fail to uncover the underlying relationship between transferable features and joint distributions. This paper proposes Learning to Match (L2M) to automatically learn the cross-domain distribution matching without relying on hand-crafted priors on the matching loss. Instead, L2M reduces the inductive bias by using a meta-network to learn the distribution matching loss in a data-driven way. L2M is a general framework that unifies task-independent and human-designed matching features. We design a novel optimization algorithm for this challenging objective with self-supervised label propagation. Experiments on public datasets substantiate the superiority of L2M over SOTA methods. Moreover, we apply L2M to transfer from pneumonia to COVID-19 chest X-ray images with remarkable performance. L2M can also be extended in other distribution matching applications where we show in a trial experiment that L2M generates more realistic and sharper MNIST samples.
\end{abstract}

\section{Introduction}

Traditional machine learning generally assumes that training and test data are from the same distribution. In reality, this i.i.d. assumption barely holds. When an algorithm is trained on one domain and then tested on another domain, the performance is likely to drop due to the different data distribution~\cite{yosinski2014transferable}. Since the collection of massive labeled data is expensive and time-consuming, a more promising approach is to perform domain adaptation (DA) to enable the consistent performance of a predictive function on different domains.

The core challenge of DA is to match the cross-domain joint distributions~\cite{ben2007analysis}. However, the labels on the target domain are often unavailable in unsupervised DA. Therefore, a trend is to approximately match the joint distributions by matching the marginal and conditional distributions as theoretically verified in~\cite{zhao2019learning,ben2007analysis}. Existing approaches achieve this goal via learning a domain-invariant representation by minimizing predefined distribution distances such as MMD~\cite{pan2011domain,long2015learning,tzeng2014deep, wang2018visual}, or an implicit discrepancy by an adversarial min-max game~\cite{ganin2014unsupervised,tzeng2017adversarial,zhang2019bridging,courty2016optimal}. Recent works suggest that in addition to jointly matching these two distributions with equal weights~\cite{long2013transfer}, an adaptive weighting scheme is necessary 
to achieve better distribution matching performance~\cite{wang2017balanced,wang2018visual,fang2019open,yu2019transfer}. 

Unfortunately, it remains challenging to apply DA to new applications. Existing methods are built with their own priors and inductive bias in approximating the joint distribution matching, which may fail to uncover the underlying relationship between transferable features and joint distributions~\cite{johansson2019support}. For instance, MMD~\cite{gretton2012kernel} may not be discriminative enough for high-dimensional data, and Jensen-Shannon divergence is not sensitive to mode collapse~\cite{ramdas2015decreasing,hu2017unifying,huszar2015not,theis2015note}. A recent work Learning to Transfer (L2T)~\cite{wei2018transfer} aims to reduce such bias by learning the ``transfer experience'' from thousands of pre-computed tasks before applying to new problems. However, L2T needs to build historical tasks from large auxiliary datasets, which is expensive and burdensome. Since deep learning makes it possible to learn features directly from the original datasets, can we design an automatic distribution matching strategy in a data-driven way?

% We still need a brutal-force exploration of existing algorithms or a heuristic selection of them for optimal performance, which requires considerable expertise in an ad-hoc and unsystematic manner~\cite{wei2018transfer}. Moreover, each method is built with its own prior and bias, making it challenging to determine the most suitable one for a given problem.  

In this work, we propose a Learning to Match (L2M) framework to automatically match the cross-domain distributions while reducing the inductive bias on matching functions. Stepping back from the hand-crafted and predefined distances, we construct a meta-network to learn the distribution matching functions directly from the source and target domains. The meta-network is an MLP network which is theoretically a universal approximator for almost any continuous function~\cite{csaji2001approximation}. We design a novel matching feature generator to L2M, where both task-independent and human-designed matching features can be taken as inputs to the meta-network for better distribution matching. Therefore, L2M can be seen as a general framework that unifies the deep features and human-crafted features (pre-defined distances) from the view of traditional vs. deep learning. Since it is challenging to optimize L2M with the unavailability of target domain labels, we propose to construct and update meta-data in a self-supervised manner~\cite{jing2020self} for updating the distribution matching loss. On the basis of matching features and meta-data, we propose an online optimization algorithm for L2M which can achieve accurate and steady performance.

%This naturally leads to rethinking domain adaptation in the view of meta-learning. In fact, meta-learning and transfer learning have long been sharing the same goal towards establishing a positive bias with training data that benefits the performance on test data~\cite{bengio1992optimization,santoro2016meta,snell2017prototypical,vinyals2016matching,finn2017model,ravi2016optimization,shu2019meta,pham2020meta}. Meta-learning solves the adaptation problem by transferring the meta-knowledge from a number of training tasks. Intuitively, it seems promising if we directly apply some successful meta-learning algorithms to DA by constructing meta-learning tasks using the source and target domains as meta-train and meta-test tasks, respectively. However, this does not work for three reasons: (1) Tasks in meta-learning are assumed i.i.d.~\cite{finn2017model}, which barely holds in domain adaptation. (2) Few-shot task sampling could lose the category information (\textit{e.g.} how to construct the $N$-way $K$-shot meta-learning tasks in a DA 
%Office-Home dataset~\cite{venkateswara2017deep} with 65 classes?). (3) There is not enough meta-knowledge to transfer from many tasks due to catastrophic forgetting~\cite{serra2018overcoming,bengio2020meta}. Therefore, a guideline is needed to benefit DA with meta-learning.

Experiments show that L2M outperforms several state-of-the-art methods on public DA datasets. L2M is a general and flexible framework that can be used in other cross-domain tasks. We apply L2M to COVID-19 X-ray image classification by transferring knowledge from normal pneumonia to COVID-19, where L2M outperforms other methods in this data-hungry and imbalanced task. As an extension, L2M can be used for generating more realistic and sharper hand-written digits. The code of L2M will be released soon at \url{https://github.com/jindongwang/transferlearning/tree/master/code/deep/Learning-to-Match}.

\section{Related Work}

\paragraph{Transfer learning and domain adaptation.}
Domain adaptation (DA) is a specific area of transfer learning~\cite{pan2010survey}. Existing works tend to explicitly or implicitly reduce the distribution divergence. The explicit distances are predefined divergence, such as Maximum Mean Discrepancy (MMD)~\cite{gretton2012kernel}, KL or JS divergence, cosine similarity, mutual information, and higher-order moments~\cite{zellinger2017central}, which are well investigated in recent DA works~\cite{pan2011domain,long2015learning,tzeng2014deep,sun2016deep,wang2018visual}. Optimal transport (OT) is another popular measure for distribution matching~\cite{courty2014domain,courty2016optimal,courty2017joint,bhushan2018deepjdot,zhang2019optimal}. There are other geometrical distances or transforms such GFK~\cite{gong2012geodesic} and subspace learning~\cite{sun2015subspace,sun2016return}. The implicit distance indirectly bridges the distribution gap through adversarial nets~\cite{goodfellow2014generative}, or learnable metrics~\cite{luo2018transfer}. GAN-based DA methods learn domain-invariant features by confusing the feature extractor and discriminator~\cite{zhang2019bridging,tzeng2017adversarial,ganin2014unsupervised}, while metric learning~\cite{luo2018transfer} focuses on the sample-wise distance. Recent research implies performance improvement by adding more prior to the matching strategy such as adaptive weights between marginal and conditional distributions~\cite{wang2018visual,fang2019open,yu2019transfer} with weights generated by the $\mathcal{A}$-distance~\cite{ben2007analysis}. Learning to transfer (L2T)~\cite{wei2018transfer} is similar to our idea in spirit. However, L2T has to manually construct thousands of transfer tasks to learn a linear transformation matrix using MMD, while L2M does not rely on historical tasks and learns non-linear feature maps, which is more efficient and general. There are several works aiming at bridging two domains by normalization such as BN~\cite{ioffe2015batch}, AutoDIAL~\cite{cariucci2017autodial}, AdaBN~\cite{li2018adaptive}, and TransNorm~\cite{wang2019transferable}, which did not focus on direct learning the cross-domain joint distributions.

\paragraph{Distribution matching.}
Generative adversarial nets (GANs)~\cite{goodfellow2014generative} matches distributions between training and generated samples by iteratively training a domain discriminator and generator to confuse the discriminator. Our L2M is model-agnostic that can be applied in an adversarial manner by adopting GAN-based schemes such as DANN~\cite{ganin2014unsupervised} or can also work without GAN. The pixel-level DA~\cite{bousmalis2017unsupervised} learns the distribution matching in pixel-space.

\section{Methodology}

%We mainly focus on solving the DA problem in classification tasks, which can easily be extended to other settings. Let $\pmb{x} \in \mathcal{X}$ and $y \in \mathcal{Y}$ denote the inputs and outputs, respectively. In UDA, we are given a source domain $\mathcal{D}_{s}=\{(\pmb{x}^{s}_{i},y^{s}_{i})\}^{n_s}_{i=1}$ of $n_{s}$ labeled examples and a target domain $\mathcal{D}_{t}=\{\pmb{x}^{t}_{j}\}^{n_t}_{j=1}$ of $n_{t}$ unlabeled examples. In DA, $\mathcal{X}_s = \mathcal{X}_t, \mathcal{Y}_s = \mathcal{Y}_t$. Since $\mathcal{D}_s \sim P_s(\pmb{x},y)$ and $\mathcal{D}_t \sim P_t(\pmb{x},y)$, the goal of deep UDA is to design a deep neural network that enables learning of transfer classifiers $h_\phi (\pmb{x}): \mathcal{X} \rightarrow \mathcal{Y}$ parametrized by $\phi \in \Phi \equiv \mathbb{R}^d$ that formally reduces the domain shifts such that the target risk $\epsilon_{t}(h) = \mathbb{E}_{(\pmb{x},y) \sim P_t}[h(\pmb{x}) \neq y]$ can be bounded by using the source domain while achieving better performance on the target domain.

\subsection{Learning to Match}
\label{sec-reformulation}
We can decompose $h_\phi$ into a feature extractor $f_\phi$ and a classification layer $G_y$, where $f_\phi$ is explicitly parameterized by $\phi$ since it is more important for domain-invariant representation learning.
Under the principle of structural risk minimization (SRM)~\cite{sugiyama2015introduction}, the optimal model parameter can be learned as:
\begin{equation}
    \phi^\star = \arg \min_\phi \mathcal{L}_{\mathrm{cls}}(G_y \circ f_\phi;\mathcal{D}_s) + \lambda \mathcal{L}_{\mathrm{match}}(f_\phi(\mathcal{D}_s), f_\phi(\mathcal{D}_t)),
\end{equation}
where $\mathcal{L}_{\mathrm{cls}}$ is the classification loss on the source domain, $\mathcal{L}_{\mathrm{match}}$ is the distribution matching loss, and $\lambda$ is a trade-off parameter.

It is challenging to directly match the cross-domain joint distributions since the labels for the target domain are not available. Therefore, existing methods tend to approximate $\mathcal{L}_{\mathrm{match}}$ using different priors. For instance, if we let $\mathcal{L}_{\mathrm{match}}=d(\mathcal{D}_s, \mathcal{D}_t)$ where $d$ is a predefined distance such as MMD~\cite{pan2011domain}, then we can get the explicit distribution matching. If $\mathcal{L}_{\mathrm{match}} = \mathbb{D}(\mathcal{D}_s, \mathcal{D}_t)$ where $\mathbb{D}$ is the adversarial discriminator~\cite{ganin2014unsupervised}, then we get the implicit distribution matching. In a nutshell, the main difference among existing works is the design of explicit or implicit $\mathcal{L}_{\mathrm{match}}$.

% This scheme should be significantly expanded so as to drop the inductive bias. An easy example is what if $\ell_{\mathrm{m}}$ and $\ell_{\mathrm{c}}$ can be integrated in a more principled way other than linear combination, or what if we can learn other high-level matching loss that does not need $\ell_{\mathrm{m}}$ and $\ell_{\mathrm{c}}$?

\begin{figure}[t!]
	\centering\includegraphics[width=\textwidth]{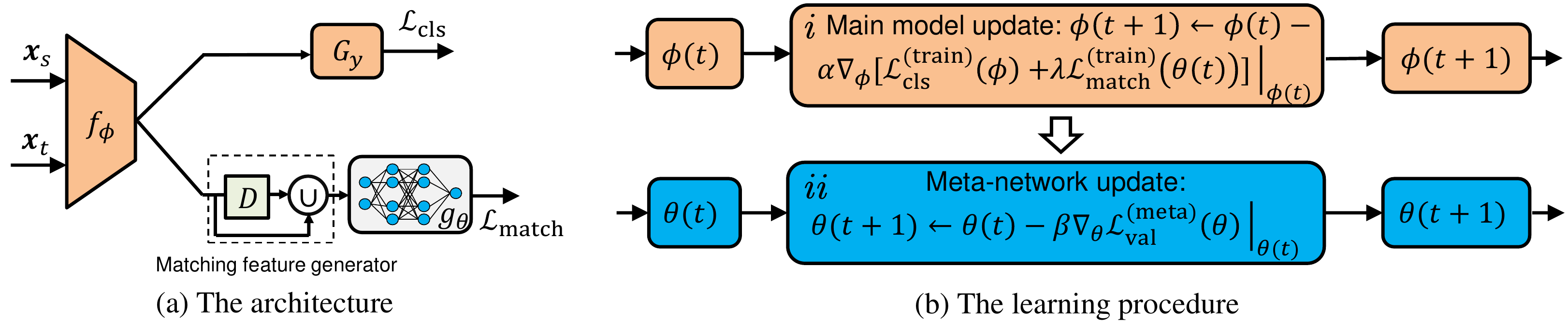}
	\vspace{-.1in}
	\caption{The framework and computing flow of the proposed L2M approach.}
	\label{fig-framework}
	\vspace{-.15in}
\end{figure}

In this paper, we postulate L2M to automatically match the distributions across domains. The core of L2M is a meta-network that learns the distribution matching in a data-driven manner. To be specific, the meta-network is a Multi-Layer Perceptron (MLP) that has the ability to approximate any continuous functions~\cite{csaji2001approximation}. Therefore, L2M can learn the distribution matching loss directly from the source and target domains:
\begin{equation}
\label{eq-gtheta}
\mathcal{L}_{\mathrm{match}} (\mathcal{D}_s, \mathcal{D}_t) = g_\theta (f_\phi(\mathcal{D}_s), f_\phi(\mathcal{D}_t)),
\end{equation}
where $g(\cdot)$ is the distribution matching function (network) parameterized by $\theta$. It is clear that this formulation is a general form that can theoretically include existing pre-defined distances.

With the meta-network $g_\theta$, we build a general framework as shown in Fig.~\ref{fig-framework}(a). The architecture consists of four parts: feature extractor $f_{\phi}$, label classifier $G_y$, meta-network $g_\theta$, and matching feature generator. Specifically, $f_{\phi}$ is a CNN network to extract the features of the input domains, $G_y$ is trained to minimize the prediction loss on the labeled (source) domain, and $g_{\theta}$ is an MLP network, which is used to match the cross-domain distributions (learn $\mathcal{L}_{\mathrm{match}}$). The most important part is the matching feature generator, which generates useful inputs to the meta-network $g_\theta$. For a general framework that allows both deep and human-designed features, we concatenate the deep features (direct link) with the human-designed distances (the green module $D$) via the concatenation module $\cup$. Then, the matching features can be taken as inputs by the meta-network $g_\theta$ to learn the distribution matching functions.

Our learning objective is well in accordance with the DA theory proposed in~\cite{ben2007analysis} that directly learns to reduce the distribution divergence between domains, such that the risk on the target domain can be bounded. The learning objective of L2M can be obtained as:
\begin{equation}
\label{eq-optim}
\min_{\theta, \phi} = \arg \min \mathcal{L}_{\mathrm{cls}}(\phi) + \lambda \mathcal{L}_{\mathrm{match}}(\phi;\theta).
\end{equation}

However, it remains challenging to optimize the above equation due to three reasons. Firstly, what kind of matching features should we take as inputs to the meta-network $g_\theta$ for better distribution matching? Secondly, we only have the labeled source domain and the unlabeled target domain, how to compute the distribution matching loss $\mathcal{L}_{\operatorname{match}}$ without the target domain labels? Thirdly, even if we have the matching features and optimization data, we cannot use a simple EM algorithm for optimization since updating $\mathcal{L}_{\operatorname{cls}}$ and $\mathcal{L}_{\operatorname{match}}$ on the same training data will definitely lead to overfitting and local optimum. Therefore, it is still non-trivial to optimize L2M.

In the next sections, we introduce how to tackle the above three challenges.

\subsection{Matching feature generator}

The matching feature generator generates useful representations as inputs to the meta-network $g$. We use $\mathbf{F}$ to denote the matching features. Technically, $\mathbf{F}$ can be any useful representations. In this paper, we propose two different kinds of matching features as shown in Table~\ref{tb-match-feat}: (1) \textbf{Task-independent features}, which are general and can be automatically computed by the main network $f_\phi$ as shown in the direct link of Fig.~\ref{fig-framework}(a); (2) \textbf{Human-designed distances (features)} as indicated in the green module $D$ in Fig.~\ref{fig-framework}(a), which are the pre-defined distances such as MMD or adversarial game. These features can either be used alone, or be concatenated by the concatenation module $\cup$. In later experiments, it is surprising to find that the combination of these two features can be seen as the combination of deep and human-designed features, which generally leads to better performance. More details of the matching features are in the supplementary file.

% \usepackage{multirow}
\begin{table}[htbp]
\centering
\caption{Description of the matching features. $q(\cdot)$ is the function of last layer before softmax. $d_\mathrm{m}$ and $d_\mathrm{c}$ are marginal and conditional explicit distances using MMD. $\mathbb{D}_\mathrm{m}$ and $\mathbb{D}_\mathrm{c}$ are marginal and conditional implicit distances using adversarial min-max game. $[a,b]$ is the concatenation of $a$ and $b$.}

\label{tb-match-feat}
\resizebox{\textwidth}{!}{
\begin{tabular}{cccc}
\toprule
Feature Type & Notation & Description & Calculation \\ \hline
\multirow{2}{*}{Task-independent} & $\mathbf{F}_{\operatorname{emb}}$ & Feature embedding & $\mathbf{F}_{\operatorname{emb}}=[f_\phi(\pmb{x}_i), f_\phi(\pmb{x}_j)]$ \\ 
 & $\mathbf{F}_{\operatorname{logit}}$ & logit & $\mathbf{F}_{\operatorname{logit}}=[q(f_\phi(\pmb{x}_i)), q(f_\phi(\pmb{x}_j))]$ \\ \hline
\multirow{2}{*}{\makecell[c]{Human-designed \\ (pre-defined dist.)} } & $\mathbf{F}_{\operatorname{mmd}}$ & Explicit distribution distance & $\mathbf{F}_{\operatorname{mmd}} = [d_\mathrm{m}(f_\phi(\pmb{x}_i),f_\phi(\pmb{x}_j)), d_\mathrm{c}(f_\phi(\pmb{x}_i),f_\phi(\pmb{x}_j))]$ \\ 
 & $\mathbf{F}_{\operatorname{adv}}$ & Implicit distribution distance & $\mathbf{F}_{\operatorname{adv}} = [\mathbb{D}_\mathrm{m}(f_\phi(\pmb{x}_i),f_\phi(\pmb{x}_j)), \mathbb{D}_\mathrm{c}(f_\phi(\pmb{x}_i),f_\phi(\pmb{x}_j))]$ \\ \bottomrule
\end{tabular}
}

\end{table}

\subsection{The construction of meta-data}
We introduce the idea of ``meta-data''.
Since direct computation of the distribution matching loss $\mathcal{L}_{\operatorname{match}}$ is hard due to the unavailability of target labels, we turn to using the meta-data $\mathcal{D}_{\mathrm{meta}}$ instead. To be more specific, $\mathcal{D}_{\mathrm{meta}}=\{\pmb{x}^{t}_{j}\}^{m \times C}_{j=1} \sim P_t(\pmb{x},\hat{y})$ where $\hat{y}$ is the predicted (pseudo) label on the target domain. In each iteration, we randomly sample $m$ instances for each class with high prediction scores calculated by the main network as the ground truth of the meta-data. This selection is iterated in the whole learning process for better performance. The pseudo labels of the meta-data can get more confident since the meta-data are chosen from the target domain data with the highest prediction probabilities. This assumption is validated in early works~\cite{zhang2019bridging,wang2018visual} and can also be seen as a self-supervised technique~\cite{jing2020self}. Therefore, the matching loss is calculated on the training data ($\mathcal{L}_{\mathrm{match}}=\mathcal{L}^{(\mathrm{train})}_{\mathrm{match}}$) when updating the main network $f_\phi$, and the meta-network $g_\theta$ is updated on the meta-data.
% \subsection{Discussions}

% Compared to existing works, L2M learns the distribution matching in a data-driven manner, instead of using heuristic matching functions with predefined distances. By taking different matching features, L2M can be more powerful. Therefore, existing works that use hand-crafted matching functions and distances can be seen as shallow learning methods, while L2M is the \textit{deep} version. 
% For instance, recent researches that learn $\mathcal{L}_{\mathrm{match}}$ by matching marginal and conditional distributions via giving equal~\cite{long2013transfer} or adaptive weights~\cite{wang2017balanced,wang2018visual,fang2019open} can be formulated as:
% \begin{equation}
% \label{eq-previous}
%     \mathcal{L}_{\mathrm{match}} = (1 - \mu) \ell_{\mathrm{m}}(\mathcal{D}_s, \mathcal{D}_t) + \mu \ell_{\mathrm{c}}(\mathcal{D}_s, \mathcal{D}_t),
% \end{equation}
% where $\mu=0$ and $\mu=1$ indicates marginal distribution matching ($\ell_{\mathrm{m}}$) and conditional distribution matching ($\ell_{\mathrm{c}}$) respectively, and $\mu \in (0,1)$ denotes a linear combination. It is clear that they are only special cases of our L2M.

% In fact, L2M is a general and model-agnostic framework that can be applied to cross-domain tasks by altering the architectures and matching features. L2M can also be used for distribution matching in generative models.

\subsection{Learning algorithm}
In this paper, we propose an online updating algorithm for L2M. Fig.~\ref{fig-framework}(b) illustrates the key learning steps. It should be noted that the data for updating $\phi$ and $\theta$ are different: when updating $\phi$, we use the normal training data from the source domain to calculate the cross-entropy loss; when updating $\theta$, we use the source domain and the pseudo-labeled target domain meta-data. The learning procedure of L2M consists of two main steps: main network update and meta-network update. In the following, we use $t$ to denote learning steps.

\paragraph{Main network update.}
This step is mainly for updating $\phi$ for the main network. To enforce the update of $\theta$ in the next step, we construct an assist model which is a copy of the main model by inheriting the same architecture and parameters from the main model ($f_\phi, G_y, g_\theta$) and use it for calculating the loss.
We employ SGD for optimizing the classification loss $\mathcal{L}_{\mathrm{cls}}$ and distribution matching loss $\mathcal{L}_{\mathrm{match}}$. $\mathcal{L}_{\mathrm{cls}}$ can be formulated as:
\begin{equation}
\label{equ-clsloss}
\mathcal{L}^{\operatorname{(train)}}_{\mathrm{cls}} = \mathbb{E}_{(\pmb{x}, y)\sim B_s} \ell^{(\text{CE})}(G_{y}(f_{\phi}(\pmb{x})), y),
\end{equation}
where $\ell^{(\text{CE})}$ is the cross-entropy loss and $B_s$ denotes a mini-batch data sampled from $\mathcal{D}_s$. The distribution matching loss $\mathcal{L}_{\mathrm{match}}$ is calculated by the meta-network $g_{\theta}$:
\begin{equation}
\label{equ-jointloss}
\begin{split}
\mathcal{L}^{\operatorname{(train)}}_{\mathrm{match}} = \mathbb{E}_{\pmb{x}_i\sim B_s,\pmb{x}_j\sim B_t} g_{\theta}(f_\phi(\pmb{x}_i),f_\phi(\pmb{x}_j);\phi),
\end{split}
\end{equation}
where $B_t$ is a mini-batch data sampled from $\mathcal{D}_t$. Note that this step does not need the meta-data from the target domain since we only sample a batch of source and target domain data ($\pmb{x}$) and do not need the target domain label $y$. Therefore, we do not update the matching loss.

%Apparently, the matching features $F(\pmb{x}_i,\pmb{x}_j;\phi)$ need to be learnt. We will introduce how to learn them explicitly and implicitly in the next section. For now, we can treat them as already-obtained components.

After getting the training loss, the updating equation of the copied main model can be obtained by moving the current $\phi(t)$ towards the descent direction of objective in Eq.~(\ref{eq-optim}):
% \begin{align}
% \label{eq-MML1}
%     \widehat{\phi}(t)~=~&\phi(t)-\alpha \mathbb{E}_{(\pmb{x}_i,y_i)\sim B_s}\nabla_\phi \ell^{(\text{CE})}(G_{y}(f_{\phi}(\pmb{x}_{i}), y_i))|_{\phi(t)}\nonumber\\
%     &+\lambda \mathcal{L}^{\operatorname{(train)}}_{\mathrm{match}}(f_\phi(\pmb{x}_i),f_\phi(\pmb{x}_j);\phi(t));\theta(t)),
% \end{align}
\begin{equation}
\label{eq-MML1}
\phi(t+1)=\phi(t)-\alpha \nabla_\phi [\mathcal{L}^{\operatorname{(train)}}_{\operatorname{cls}}(\phi) + \lambda \mathcal{L}_{\operatorname{match}}^{\operatorname{(train)}}(\phi; \theta(t))] | _{\phi(t)},
\end{equation}
where $\alpha$ is the learning rate of the assist model. $\mathcal{L}^{\operatorname{(train)}}_{\operatorname{cls}}(\phi)$ and $\mathcal{L}^{\operatorname{(train)}}_{\operatorname{match}}(\phi; \theta(t))$ are computed by Eq.~(\ref{equ-clsloss}) and (\ref{equ-jointloss}).

\paragraph{Meta-network update.}
This step is for updating $\theta$ for the meta-network $g_\theta$ on the meta-data $\mathcal{D}_{\mathrm{meta}}$. Similar to updating $\phi$, it is natural that updating $\theta$ requires ``ground-truth'' available for the distribution matching loss $\mathcal{L}_{\operatorname{match}}$. However, this is not available in UDA problems. To solve this challenge, we employ a self-supervised strategy with the assumption that after one epoch of updating $\phi(t)$ to $\phi(t+1)$, the distribution matching loss can get smaller with the increasing confidence of the target pseudo labels. This pseudo-label assumption is widely adopted in previous DA works~\cite{zhang2019bridging, wang2018visual,tzeng2014deep}. Therefore, this validation loss can be updated by computing the discrepancy between the distribution matching loss on $\phi(t)$ and $\phi(t+1)$:
\begin{equation}
\label{equ-valloss}
\mathcal{L}^{(\operatorname{meta})}_{\operatorname{val}} = \mathbb{E}_{\pmb{x}\sim\mathcal{D}_{\mathrm{meta}}} \tanh(g_\theta(f(\pmb{x};\phi(t)))-g_\theta(f(\pmb{x};\phi(t+1)))),
\end{equation}
where $\tanh(\cdot)$ is an activation function. Note that we fix $\phi$ in this step and minimize Eq.~(\ref{equ-valloss}) w.r.t. $\theta$ can gradually update the meta-network $g_\theta$. The pseudo labels can be easily obtained by a single forward-pass and then selected according to the confidence (softmax probability). To ensure their confidence, we choose the samples with probabilities $\ge 0.8$ in our experiments.

Denote $\beta$ the learning rate of meta-network $g_\theta$, then $\theta$ can be updated as:
\begin{equation}
\label{eq-theta}
\theta(t+1) = \theta(t)-\beta \nabla_{\theta} \mathcal{L}^{(\operatorname{meta})}_{\operatorname{val}}(\theta;\phi(t),\phi(t+1)) | _{\theta(t)}.
\end{equation}

The above two steps are used iteratively as the pseudo labels of the meta-data can be more confident and all the losses can be iteratively minimized. In our experiments, we observe that the network will converge in dozens of epochs. The complete algorithm and convergence analysis are presented in the supplementary file.

As for inference, L2M is the same as existing DA methods~\cite{long2018conditional,wang2018visual,zhang2019bridging,tzeng2014deep}. We simply fix $\phi$ and $\theta$ and use the main model to perform a single forward-pass to get the results for the test data.

\section{Experiments on Public Datasets}
%In this section, we evaluate L2M on UDA tasks. As an extension, we show in the last experiment that L2M can also be used for image generation on MNIST dataset.

\subsection{Experimental setup}

\paragraph{Datasets.}
We adopt four public datasets: ImageCLEF-DA~\cite{imageclef}, Office-Home~\cite{venkateswara2017deep}, VisDA-2017~\cite{visda2017} and Office-31~\cite{saenko2010adapting}. They are widely used by most UDA approaches~\cite{long2018conditional,zhang2019bridging,wang2018visual,tzeng2017adversarial}. The detailed dataset descriptions are presented in the supplementary file.

\paragraph{Baselines and implementations.}
We comapre L2M with several recent DA methods: \textbf{ResNet}~\cite{he2016deep}, \textbf{DDC}~\cite{tzeng2014deep}, \textbf{DAN}~\cite{long2015learning}, \textbf{DANN}~\cite{ganin2014unsupervised}, \textbf{JAN}~\cite{long2016deep}, \textbf{MADA}~\cite{pei2018multi}, \textbf{CAN}~\cite{zhang2018collaborative}, \textbf{MEDA}~\cite{wang2018visual}, \textbf{DAAN}~\cite{yu2019transfer}, \textbf{CDAN}~\cite{long2018conditional}, \textbf{DeepJDOT}~\cite{bhushan2018deepjdot}, \textbf{MDD} ~\cite{zhang2019bridging}, and
\textbf{TransNorm}~\cite{wang2019transferable}. The main network of all methods including L2M are based on ImageNet-pretrained ResNet50. The hyperparameter setting for L2M are presented in the supplementary file. We follow the standard protocols for UDA and take classification accuracy on the target domain as the evaluation metric and target labels are only used for evaluation. The best parameters are tuned according to~\cite{you2019towards}. The results are the average accuracy of 10 experiments by following the same protocol~\cite{zhang2019bridging,long2018conditional,wang2018visual,tzeng2014deep}.

\subsection{Analysis of matching features}
Before using L2M, a natural question is which matching feature should be used for better performance. Moreover, how is the performance of MMD and adversarial discrepancy in L2M compared to existing MMD or adversarial-based DA methods? To answer these questions, we randomly choose two pairs of DA tasks from Office-Home dataset (R $\rightarrow$ A, R $\rightarrow$ P, and vice versa) to compare the performance of existing distance-based methods (DAN~\cite{long2015learning} and MEDA~\cite{wang2018visual} use MMD while DANN~\cite{ganin2014unsupervised}, CDAN~\cite{long2018conditional}, and DAAN~\cite{yu2019transfer} use adversarial-based discrepancy) with L2M.  Technically, all matching features can be combined, which will result in $2^4=16$ different matching features. For computational issue, we construct eight matching features: $\{\mathbf{F}_{\operatorname{emb}}, \mathbf{F}_{\operatorname{logit}}, \mathbf{F}_{\operatorname{mmd}}, \mathbf{F}_{\operatorname{adv}}, [\mathbf{F}_{\operatorname{emb}}, \mathbf{F}_{\operatorname{mmd}}], [\mathbf{F}_{\operatorname{emb}}, \mathbf{F}_{\operatorname{adv}}], [\mathbf{F}_{\operatorname{logit}}, \mathbf{F}_{\operatorname{mmd}}], [\mathbf{F}_{\operatorname{logit}}, \mathbf{F}_{\operatorname{adv}}]\}$. It should be noted that both $\mathbf{F}_{\operatorname{emb}}$ and $\mathbf{F}_{\operatorname{logit}}$ can be applied to both explicit (deep) and implicit (adversarial) matching networks, leading to ten features in total. In addition, we do not combine three or four features since their performance can naturally be better but with more computations.

%% Ablation experiments of L2M
\begin{table}[ht!]
	\caption{\upshape Matching features of L2M.}
	\vspace{-.1in}
	\label{tb-ablation}
	\centering 
	\resizebox{.7\textwidth}{!}{
		\begin{tabular}{c|ccccc}
			\toprule
			& Method & R$\rightarrow$A & A$\rightarrow$R & P$\rightarrow$R & R$\rightarrow$P\\ \hline
			\multirow{7}{*}{Explicit} & DAN~(marginal)~\cite{long2015learning} & 63.1 & 67.9 & 67.7 & 74.3 \\
			& MEDA (joint)~\cite{wang2018visual} & 61.2 & 68.8 & 72.9 & 76.0 \\ 
			& L2M (emb) & 71.1 & 76.1 & 79.1 & 83.7  \\
			& L2M (logit) & 70.3 & 76.6 & 79.4 & 83.6  \\
			& L2M (mmd) & 69.3 & 73.4 & 75.2 & 83.2  \\
			& L2M (emb+mmd) & 71.1 & 76.9 & 78.6 & 83.1  \\
			& L2M (logit+mmd) & 71.5 & 76.7 & 78.5 & 82.8  \\
			\hline
			\multirow{8}{*}{Implicit} & DANN (marginal)~\cite{ganin2014unsupervised} & 63.2 & 70.1 & 76.8 & 68.5 \\ 
			& DAAN (joint)~\cite{yu2019transfer} & 66.3 & 73.7 & 74.0 & 78.8 \\
			& CDAN (conditional)~\cite{long2018conditional} & 70.9 & 76.0 & 77.3 & 81.6  \\   
			& L2M (emb) & 70.8 & 77.8 & 79.3 & 83.2  \\ 
			& L2M (logit) & 71.6 & 71.7 & 79.4 & 83.6  \\
			& L2M (adv) & 72.7 & 78.5 & 80.3 & 83.1  \\ 
			& L2M (emb+adv) & 71.8 & 79.3 & 80.6 & 83.5  \\ 
			& L2M (logit+adv) & 71.4 & 76.6 & 78.6 & 82.8 \\ \bottomrule
		\end{tabular}
	}
	\vspace{-.1in}
\end{table}

\begin{figure}[t!]
	\centering
	\subfigure[Matching features]{
		\centering
		\includegraphics[width=0.23\textwidth]{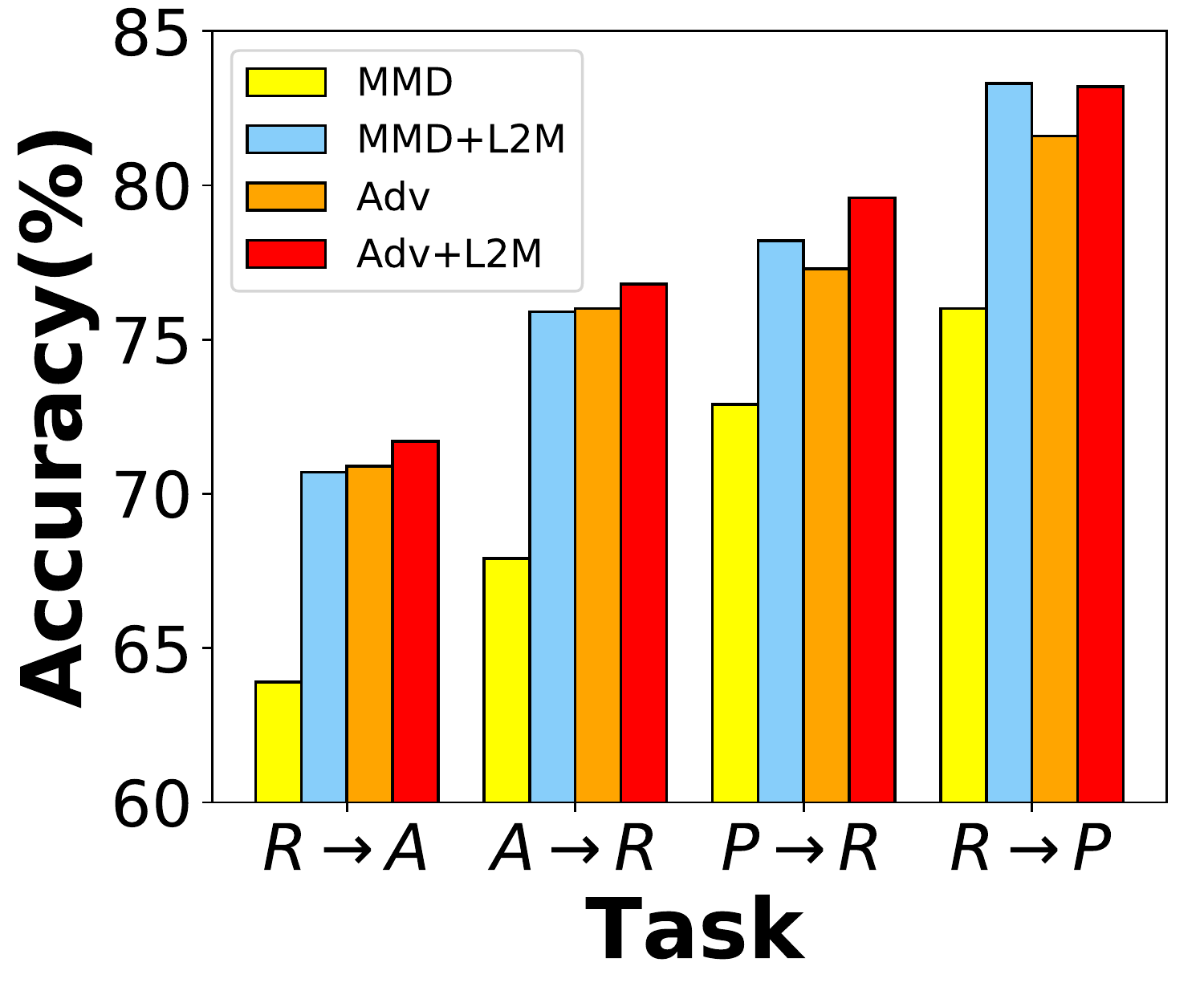}
		\label{fig-ablation-matchfea}}
	\subfigure[Meta-data]{
		\centering
		\includegraphics[width=0.23\textwidth]{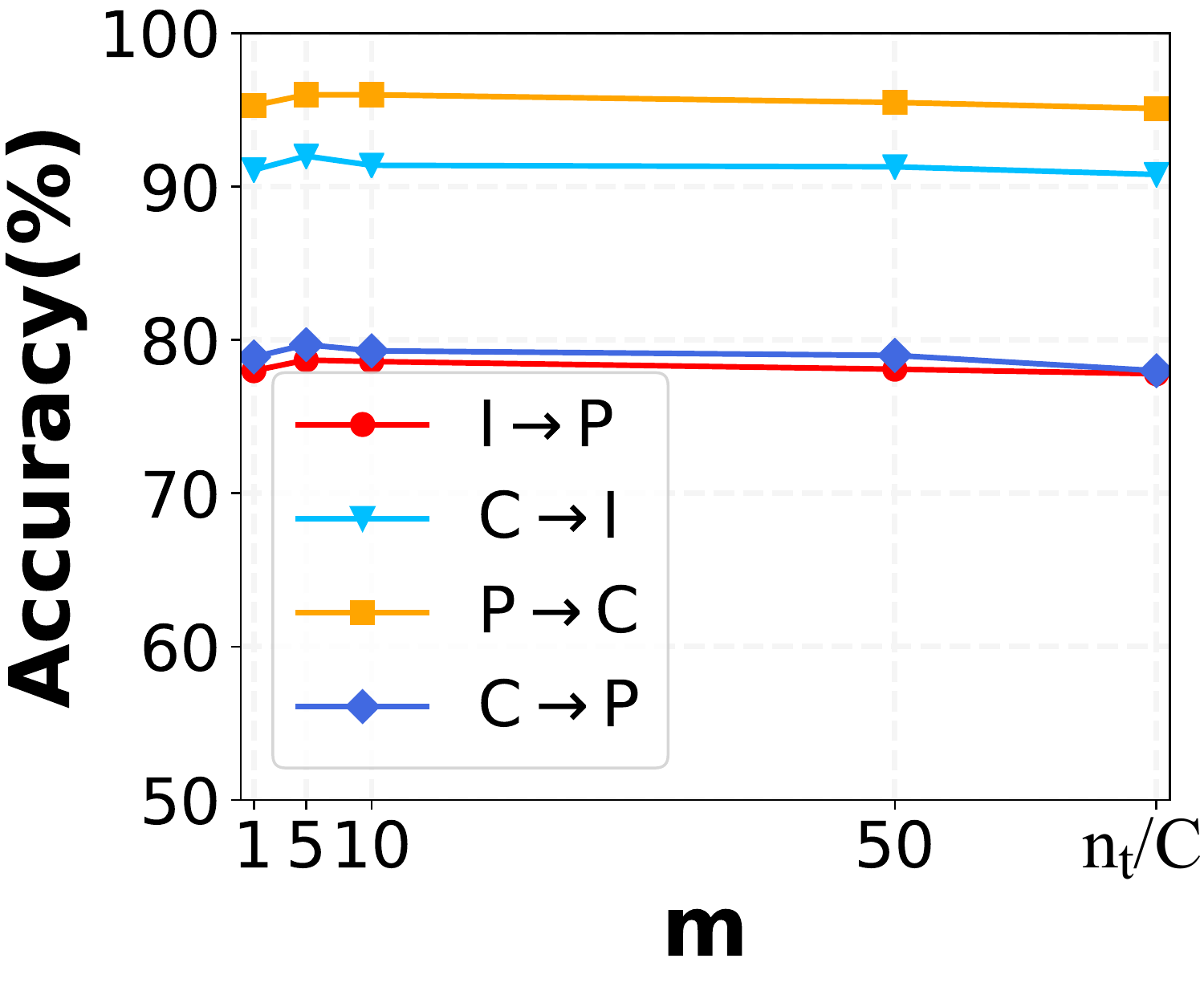}
		\label{fig-ablation-m}}
	\subfigure[Distribution distance]{
		\centering
		\includegraphics[width=0.23\textwidth]{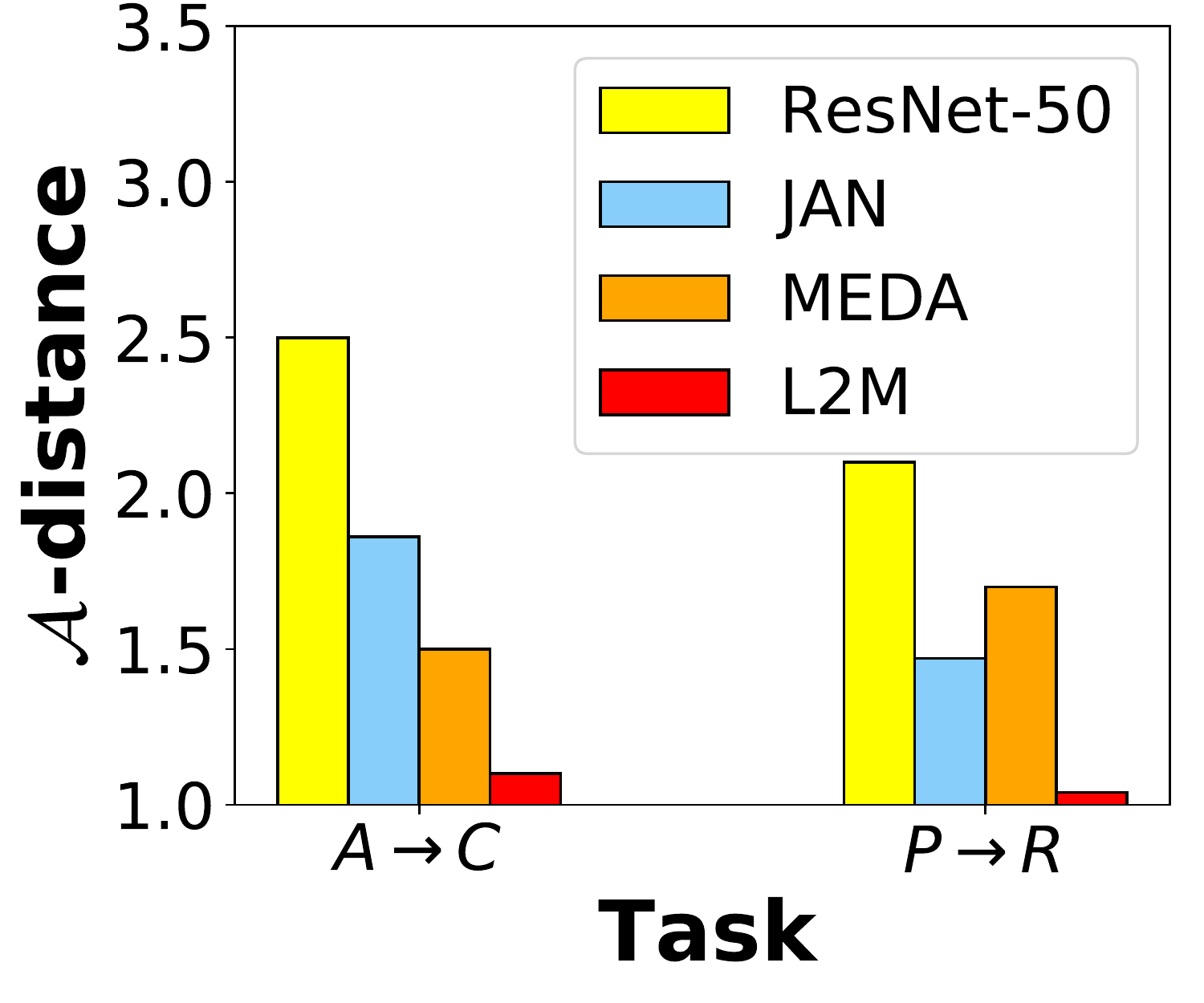}
		\label{fig-sub-adist}}
	\subfigure[Convergence]{
		\centering
		\includegraphics[width=0.23\textwidth]{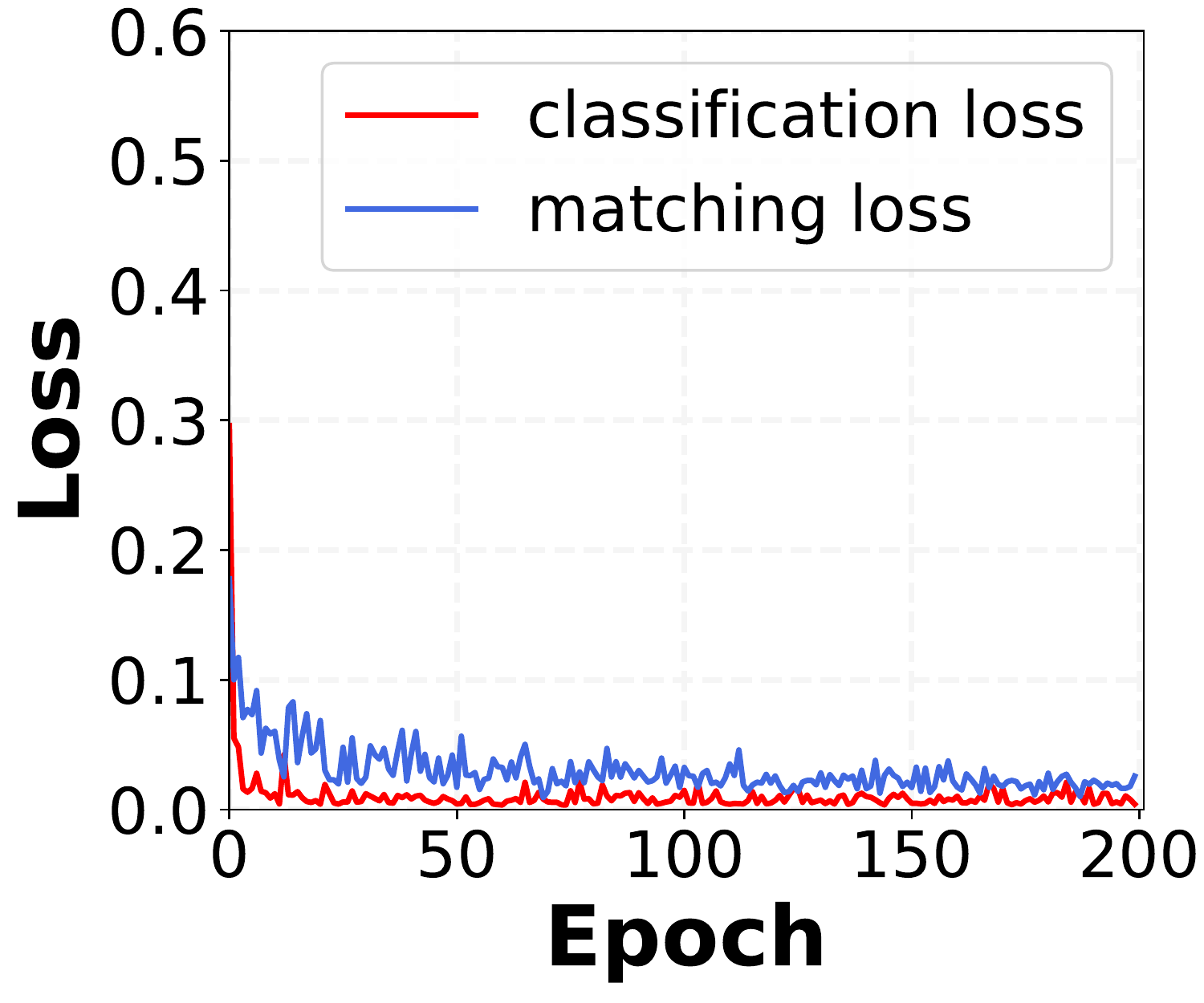}
		\label{fig-sub-loss}}
	\vspace{-.1in}
	\caption{(a) Comparison between the best existing methods with predefined distance and the average of L2M. (b) Analysis of the number of meta-data $m$. (c) Distribution discrepancy between two domains. (d) Convergence of L2M.}
	\vspace{-.1in}
	\label{fig-adist-loss}
\end{figure}

The feature dimensions of each matching feature are presented in the supplementary file. The comparison results are in Table~\ref{tb-ablation}. For better clarification, we compare the performance of best MMD- and adversarial-based methods in Fig.~\ref{fig-ablation-matchfea}, along with the \textit{average} performance of L2M using these features. More experiments can be found at the supplementary. Firstly, we see that in both explicit and implicit distribution matching, L2M can generally achieve competitive performance with different matching features. This verifies that L2M is effective for distribution matching. Secondly, in some cases, the performance of L2M with MMD distances are better than previous adversarial-based methods. Since adversarial-based methods require much more training time, this makes L2M+MMD suitable solutions for resource-constrained applications. Thirdly, the performance of L2M with both task-independent features and pre-defined distances are generally better than using each feature solely, indicating the common practice is useful that deep learning performance can be boosted by combining deep features ($\mathbf{F}_{\operatorname{emb}}$ or $\mathbf{F}_{\operatorname{logit}}$) with human-designed features ($\mathbf{F}_{\operatorname{mmd}}$ or $\mathbf{F}_{\operatorname{adv}}$). We also observe that L2M with embeddings are generally better than logits, which is probably because those embeddings contain richer information than logits. Therefore, in the next experiments, we adopt $[\mathbf{F}_{\operatorname{emb}},\mathbf{F}_{\operatorname{adv}}]$ for a balance between computation and better performance. In real applications, more domain-dependent matching features can be added according to domain knowledge.

\subsection{Performance against SOTA methods}
The results on Office-Home dataset are shown in Table~\ref{tb-officehome}, while the results on ImageCLEF-DA and VisDA-17 datasets are in Table~\ref{tb-clef}. The results on Office-31 are provided in supplementary file due to space limits. From the results, we see that the L2M outperforms all comparison methods. Specifically, on ImageCLEF-DA dataset, although the baseline for this dataset is very high, L2M still achieves an average accuracy of 89.1\% with a 0.6\% improvement over the second-best baseline. On Office-Home dataset, L2M achieves an average accuracy of 69.6\% with a 1.5\% improvement compared to the second-best. Office-Home dataset is rather complicated and involves more samples and categories, which indicates the effectiveness of L2M. On Office-31 dataset, L2M achieves an average accuracy of 89.5\%, which is also highly competitive. Last, on the VisDA-17 dataset, which is rather larger compared to the other datasets (280,000+ images), L2M achieves an accuracy of 77.5\% with a significant improvement of $\mathbf{2.9}$\%. All these results demonstrate that L2M can achieve competitive performance on DA tasks.

%% OfficeHome
\begin{table}[t!]
	\caption{\upshape Accuracy~(\%) on Office-Home for UDA (ResNet-50).}
	\vspace{-.1in}
	\label{tb-officehome}
	\centering
	\setlength{\tabcolsep}{1.0mm}{
		\resizebox{\textwidth}{!}{
			\begin{tabular}{lccccccccccccc}
				\toprule
				Method & A$\rightarrow$C  & A$\rightarrow$P  & A$\rightarrow$R  & C$\rightarrow$A  & C$\rightarrow$P  & C$\rightarrow$R  & P$\rightarrow$A  & P$\rightarrow$C  & P$\rightarrow$R  & R$\rightarrow$A  & R$\rightarrow$C  & R$\rightarrow$P  & AVG  \\ \hline
				ResNet~\cite{he2016deep} & 34.9 & 50.0 & 58.0 & 37.4 & 41.9 & 46.2 & 38.5 & 31.2 & 60.4 & 53.9 & 41.2 & 59.9 & 46.1 \\
				DAN~\cite{long2015learning}    & 43.6 & 57.0 & 67.9 & 45.8 & 56.5 & 60.4 & 44.0 & 43.6 & 67.7 & 63.1 & 51.5 & 74.3 & 56.3 \\
				DANN~\cite{ganin2014unsupervised}   & 45.6 & 59.3 & 70.1 & 47.0 & 58.5 & 60.9 & 46.1 & 43.7 & 68.5 & 63.2 & 51.8 & 76.8 & 57.6 \\
				JAN~\cite{long2016deep}    & 45.9 & 61.2 & 68.9 & 50.4 & 59.7 & 61.0 & 45.8 & 43.4 & 70.3 & 63.9 & 52.4 & 76.8 & 58.3 \\
				MEDA~\cite{wang2018visual}   & 46.6 & 68.9 & 68.8 & 49.0 & 66.4 & 66.1 & 51.8 & 45.0 & 72.9 & 61.2 & 50.3 & 76.0 & 60.2 \\
				DAAN~\cite{yu2019transfer}   & 50.5 & 65.0 & 73.7 & 53.7 & 62.7 & 64.6 & 53.5 & 45.2 & 74.0 & 66.3 & 54.0 & 78.8 & 61.8 \\
%				DSR~\cite{cai2019learning}   & 53.4 & 71.6 & 77.4 & 57.1 & 66.8 & 69.3 & 56.7 & 49.2 & 75.7 & 68.0 & 54.0 & 79.5 & 64.9 \\
				CDAN~\cite{long2018conditional} & 50.7 & 70.6 & 76.0 & 57.6 & 70.0 & 70.0 & 57.4 & 50.9 & 77.3 & 70.9 & 56.7 & 81.6 & 65.8 \\
				ALDA~\cite{chen2020adversarial} & 53.7 & 70.1 & 76.4 & 60.2 & 72.6 & 71.5 & 56.8 & 51.9 & 77.1 & 70.2 & 56.3 & 82.1 & 66.6 \\
				CDAN+TransNorm~\cite{wang2019transferable} & 50.2 & 71.4 & 77.4 & 59.3 & 72.7 & 73.1 & 61.0 & 53.1 & 79.5 & 71.9 & 59.0 & 82.9 & 67.6 \\
				MDD~\cite{zhang2019bridging} & 54.9 & 73.7 & 77.8 & 60.0 & 71.4 & 71.8 & 61.2 & 53.6 & 78.1 & \textbf{72.5} & \textbf{60.2} & 82.3 & 68.1 \\ \hline
				L2M   & \textbf{57.5} & \textbf{74.0} & \textbf{78.5} & \textbf{63.0} & \textbf{73.1} & \textbf{72.5} & \textbf{63.5} & \textbf{56.6} & \textbf{80.5} & 72.0 & \textbf{60.2} & \textbf{83.6} & \textbf{69.6} \\ \bottomrule
			\end{tabular}
	}}
\end{table}

%% ImageCLEF
\begin{table}[t!]
	\caption{\upshape Accuracy~(\%) on ImageCLEF-DA and VisDA-2017 for UDA (ResNet-50).}
	\label{tb-clef}
	\vspace{-.1in}
	\centering 
	\setlength{\tabcolsep}{2.mm}{
		\resizebox{\textwidth}{!}{
			\begin{tabular}{lccccccc|lc}
				\toprule
				\multicolumn{8}{c|}{ImageCLEF-DA} & \multicolumn{2}{c}{VisDA-2017} \\ %\cline{2-8} \cline{10-10}
				Method  & I$\rightarrow$P & P$\rightarrow$I & I$\rightarrow$C & C$\rightarrow$I & C$\rightarrow$P & P$\rightarrow$C & AVG & Method & Syn$\rightarrow$Real \\ \hline
				ResNet~\cite{he2016deep}  & 74.8 & 83.9 & 91.5 & 78.0 & 65.5 & 91.2 & 80.7 & DAN~\cite{long2015learning} & 49.8 \\
				%		DDC~\cite{tzeng2014deep}     & 74.6 & 85.7 & 91.1 & 82.3 & 68.3 & 88.8 & 81.8  \\
				DAN~\cite{long2015learning}     & 75.0 & 86.2 & 93.3 & 84.1 & 69.8 & 91.3 & 83.3 & JAN~\cite{long2016deep} & 61.6 \\
				DANN~\cite{ganin2014unsupervised}    & 75.0 & 86.0 & 96.2 & 87.0 & 74.3 & 91.5 & 85.0 & DANN+TransNorm~\cite{wang2019transferable} & 66.3 \\
				% 		D-CORAL~\cite{sun2016deep} & 76.9 & 88.5 & 93.6 & 86.8 & 74.0 & 91.6 & 85.2 \\
				MEDA~\cite{wang2018visual}    & 78.1 & 90.4 & 93.1 & 86.4 & 73.2 & 91.7 & 85.5 & DeepJDOT~\cite{bhushan2018deepjdot} & 66.9 \\
				CAN~\cite{zhang2018collaborative}     & 78.2 & 87.5 & 94.2 & 89.5 & 75.8 & 89.2 & 85.7 & MCD~\cite{saito2018maximum} & 69.2 \\ 
				JAN~\cite{long2016deep}     & 76.8 & 88.0 & 94.7 & 89.5 & 74.2 & 91.7 & 85.8 & GTA~\cite{sankaranarayanan2018generate} & 69.5 \\
				MADA~\cite{pei2018multi}    & 75.0 & 87.9 & 96.0 & 88.8 & 75.2 & 92.2 & 85.8 & CDAN~\cite{long2018conditional} & 70.0 \\
				CDAN~\cite{long2018conditional} & 77.7 & 90.7 & \textbf{97.7} & 91.3 & 74.2 & 94.3 & 87.7 & CDAN+TransNorm~\cite{wang2019transferable} & 71.4 \\
				CDAN+TransNorm~\cite{wang2019transferable} & 78.3 & 90.8 & 96.7 & \textbf{92.3} & 78.0 & 94.8 & 88.5 & MDD~\cite{zhang2019bridging} & 74.6 \\ \hline
				L2M & \textbf{78.7} & \textbf{91.0} & 97.0 & 92.0 & \textbf{79.7} & \textbf{96.0} & \textbf{89.1} & L2M & \textbf{77.5} \\ \bottomrule
			\end{tabular}
	}}
\end{table}

%Combining with meta-learning and few-shot learning, we can see that DA tasks would require sufficient knowledge transfer between domains, while the traditional meta-learning setting struggles for DA tasks. The results are also in line with some recent works on meta-learning: simply finetuning the ResNet50 network will lead to bad performance, while this simple pretrain+finetune strategy works for few-shot setting~\cite{dhillon2019baseline}.

%% Analysis of meta-data
% \begin{table}[ht!]
%     \caption{\upshape Analysis of meta-data.}
% 	\label{tb-param-m}
% 	\centering 
% 	\resizebox{0.4\textwidth}{!}{%
% 	\begin{tabular}{cccccc}
% 	\toprule
%     $m$ & $1$ & $5$ & $10$ & $50$ & $n_t/C$ \\ \hline
%     I$\rightarrow$P & 78.0 & 78.7 & 78.6 & 78.1 & 77.8 \\
%     C$\rightarrow$I & 91.1 & 92.0 & 91.4 & 91.3 & 90.8 \\
%     P$\rightarrow$C & 95.3 & 96.0 & 96.0 & 95.5 & 95.1 \\
%     C$\rightarrow$P & 78.9 & 79.7 & 79.3 & 79.0 & 78.0 \\
%     \bottomrule
%     \end{tabular}%
% 	}
% \end{table}

%\begin{figure}[t!]
%	\centering\includegraphics[scale=0.35]{figures/ablation_m.pdf}
%	\vspace{-.1in}
%	\caption{Analysis of meta-data.}
%	\label{fig-ablation-m}
%	\vspace{-.1in}
%\end{figure}

\subsection{Detailed analysis}
% \paragraph{The effectiveness of matching features}
% We conduct ablation experiments to analyze L2M. For explicit matching, DAN~\cite{long2015learning} and MEDA~\cite{wang2018visual} are special cases of L2M. For implicit matching, DAAN~\cite{yu2019transfer}, CDAN~\cite{long2018conditional} and MDD~\cite{zhang2019bridging} are special cases. We further compose 4 variants of L2M using 4 different matching features (emb., logits, emb.+distance, logits+distance) as input of the meta-learner $g_\theta$, respectively. Compared to deep features (emb. and logits), the distance here denotes the human-designed MMD loss. From Table~\ref{tb-im-ex}, we can see that L2M achieves the best performance, implying the effectiveness of the learning-to-learn strategy on matching functions and L2M is able to fit a wide range of matching functions.

\paragraph{Analysis of meta-data.}
We empirically analyze the batch size $m$ of the meta-data $\mathcal{D}_{\mathrm{meta}}$. It is obvious that a larger $m$ will bring more uncertainty, and a smaller $m$ is likely to make the meta-network unstable. We record the performance of L2M using different values of $m$ on several randomly selected tasks in Fig.~\ref{fig-ablation-m}. The results indicate that L2M is robust to $m$ and a small $m$ can lead to competitive performance. Therefore, we set $m=5$ in our experiments for computational efficiency.

\paragraph{Distribution discrepancy.}
The $\mathcal{A}$-distance~\cite{ben2007analysis} measures the distribution discrepancy that is defined as $d_{\mathcal{A}}=2(1-2\epsilon)$, where $\epsilon$ is the classifier loss to discriminate the source and target domains. Smaller $\mathcal{A}$-distance indicates better domain-invariant features. Fig.~\ref{fig-sub-adist} shows that L2M can achieve a lower $d_{\mathcal{A}}$, implying a lower generalization error of L2M.

%% Covid
\begin{table}[t!]
	\caption{\upshape Results on COVID-19 X-ray adaptation (normal pneumonia $\rightarrow$ COVID-19, ResNet-18). Here we use the 95\% confidence interval, where the corresponding value of $z$ is 1.96. The computed confidence interval $r$ is around 1.3\%.}
	\label{tb-covid}
	\vspace{-.1in}
	\centering
	\setlength{\tabcolsep}{5.0mm}{
		\resizebox{0.7\textwidth}{!}{
		\begin{tabular}{lccc}
		\toprule
        Method & Precision (\%) & Recall (\%) & F1 (\%) \\ \hline
        Train on source & 63.5 & 66.7 & 65.0 \\
        Train on target \textit{(ideal state)} & \textbf{91.7} & 55.0 & 68.8 \\
        Fine-tuning & 56.3 & 75.0 & 64.3 \\
        DLAD~\cite{zhang2020covid} & 62.0 & 73.3 & 67.2 \\
        DANN~\cite{ganin2014unsupervised} & 61.4 & 71.7 & 66.2 \\
        MCD~\cite{saito2018maximum} & 63.2 & 60.0 & 61.5 \\
        CDAN+TransNorm~\cite{wang2019transferable} & 85.0 & 39.2 & 63.7 \\ \hline
        L2M & 70.1 & \textbf{78.3} & \textbf{74.0} \\
        \bottomrule
        \end{tabular}
	}}
	\vspace{-.1in}
\end{table}

\paragraph{Convergence analysis.}
L2M introduces a meta-network, which may make the training process harder. In this section, we empirically evaluate the convergence of L2M. As shown in Fig.~\ref{fig-sub-loss}, the results on a randomly-chosen task show that L2M can reach a quick and steady convergence in a limited number of iterations. Therefore, L2M can be easily trained.

\section{Application to COVID-19 Chest X-ray Image Classification}
Other than public datasets, we apply L2M to a COVID-19 chest X-ray image classification dataset~\cite{zhang2020covidda}, where the source domain is normal or pneumonia, and the target domain is normal or COVID-19 pneumonia. Note that this is a class-imbalanced task which is more challenging and realistic. We use F1, Precision, and Recall as the evaluation metrics for this highly-imbalanced binary classification task. As shown in Table~\ref{tb-covid}, L2M achieves better results compared to finetune and other DA methods. Here we use the 95\% confidence interval, where the corresponding value of $z$ is 1.96. The computed confidence interval $r$ is around 1.3\%. More details about the dataset, comparison methods, and results are in the supplementary file.

\begin{figure}[t!]
	\centering
	\subfigure[GAN]{
		\centering
		\includegraphics[width=.3\textwidth]{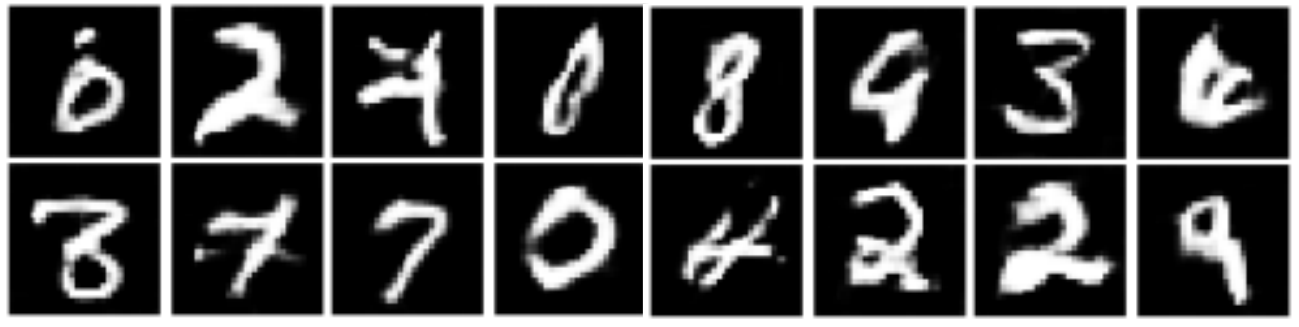}
		\label{fig-sub-gmmn}}
	\subfigure[GMMN]{
		\centering
		\includegraphics[width=.3\textwidth]{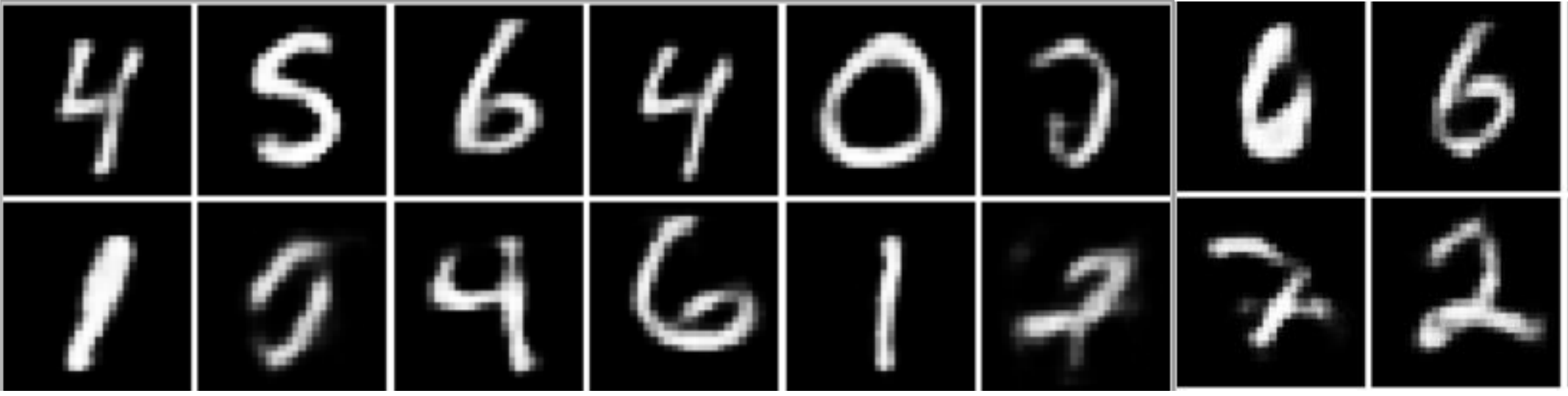}
		\label{fig-sub-gmmn}}
	\subfigure[L2M]{
		\centering
		\includegraphics[width=.3\textwidth]{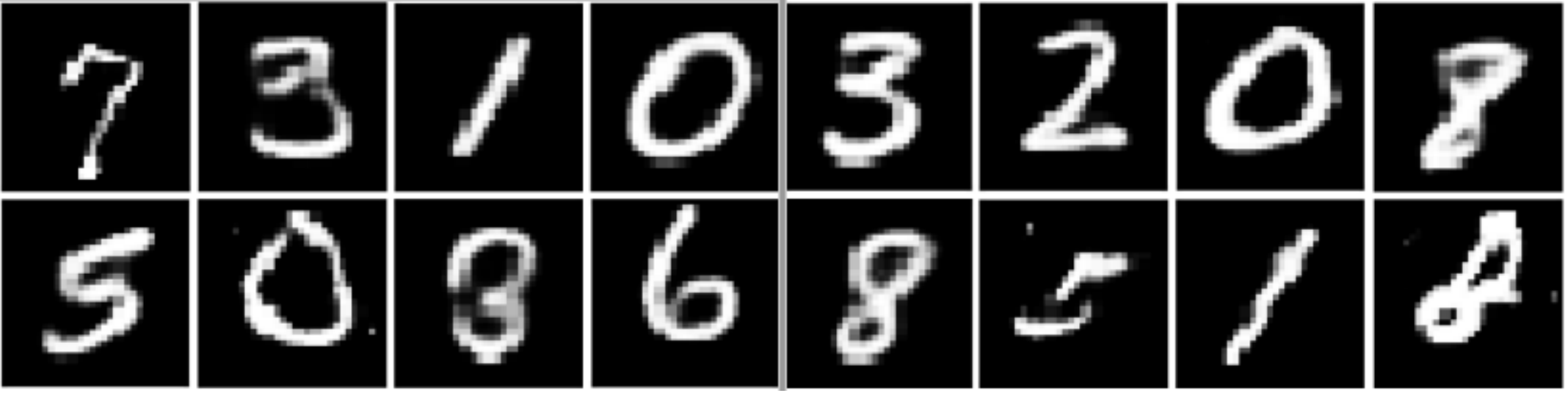}
		\label{fig-sub-l2m}}
	\vspace{-.1in}
	\caption{Generated samples from GAN, GMMN, and L2M.}
	\vspace{-.1in}
	\label{fig-generative}
\end{figure}

\section{Discussions}

\paragraph{Extending L2M for image generation.}
We show the potential of L2M in generating MNIST hand-written digits. We use GAN~\cite{goodfellow2014generative} (adversarial distance) and GMMN~\cite{li2015generative} (MMD distance) as the baselines. We replace the MMD module in GMMN with L2M. Hyperparameter settings and training details are in supplementary. The generated samples are shown in Fig.~\ref{fig-generative}. L2M can generate more realistic samples compared to GAN, and sharper samples compared to GMMN. This indicates the potential of L2M in image generation. It should be noted that this is only a \textit{trial} experiment and more efforts are needed for achieving SOTA performance on image generation.

\paragraph{Limitations and solutions.}
L2M can be roughly regarded as that requires updating two networks iteratively. Therefore, compared with regular DA methods (\textit{e.g.}, DANN, CDAN, MDD), L2M needs more than more training time. It is suggested to use a smaller batch size of meta-data compared to training data to reduce GPU memory increment and speed up training. However, the inference time is the same as other methods for using the same backbone. L2M can be more efficient by adopting knowledge distillation as suggested in meta-pseudo-labels~(MPL)~\cite{pham2020meta}, which is left for future research. Additionally, a pre-trained L2M model can be deployed to the edge devices which can achieve accurate and fast inference.

\section{Conclusions}
In this paper, for the first time, we step back from focusing on designing distribution matching features according to human knowledge, and instead, propose L2M to automatically match the cross-domain joint distributions for domain adaptation. Our work shows that by taking diverse matching features including task-independent and human-designed distances, L2M can directly learn the distribution matching in a data-driven way. L2M can be seen as a general framework that unifies deep feature learning and human-designed feature learning for better distribution matching.
Experiments on public datasets substantiate the superiority of L2M over state-of-the-art approaches on DA and image generation tasks. We apply L2M to COVID-19 X-ray image adaptation experiment, where it significantly outperforms existing methods in such a highly imbalanced task. We believe that L2M can be helpful in other problems such as domain generalization, open-set DA, and partial transfer learning, which will be the focus of future research.

%\section*{Broader Impact}
%This work mainly deals with the domain adaptation task. Therefore, We believe that for these tasks, this work will provide a positive impact on the research community. However, as indicated in the discussion section, our work also has some potential to be used for image generation tasks, e.g., generating fake images or faces, which could be a double-edged sword. Luckily, our work for image generation is only at its early stage and more future work should be done for quality refinement. For further potential risks, we will pay special attention to our source code release and the licenses to make sure that this technology will not be improperly used.

\bibliographystyle{unsrt}
\bibliography{nips20}

\begin{thebibliography}{10}

\bibitem{yosinski2014transferable}
Jason Yosinski, Jeff Clune, Yoshua Bengio, and Hod Lipson.
\newblock How transferable are features in deep neural networks?
\newblock In {\em NIPS}, pages 3320--3328, 2014.

\bibitem{ben2007analysis}
Shai Ben-David, John Blitzer, Koby Crammer, and Fernando Pereira.
\newblock Analysis of representations for domain adaptation.
\newblock In {\em NIPS}, pages 137--144, 2007.

\bibitem{zhao2019learning}
Han Zhao, Remi Tachet~des Combes, Kun Zhang, and Geoffrey~J Gordon.
\newblock On learning invariant representation for domain adaptation.
\newblock In {\em ICML}, 2019.

\bibitem{pan2011domain}
Sinno~Jialin Pan, Ivor~W Tsang, James~T Kwok, and Qiang Yang.
\newblock Domain adaptation via transfer component analysis.
\newblock {\em IEEE TNN}, 22(2):199--210, 2011.

\bibitem{long2015learning}
Mingsheng Long, Yue Cao, Jianmin Wang, and Michael~I Jordan.
\newblock Learning transferable features with deep adaptation networks.
\newblock In {\em ICML}, 2015.

\bibitem{tzeng2014deep}
Eric Tzeng, Judy Hoffman, Ning Zhang, Kate Saenko, and Trevor Darrell.
\newblock Deep domain confusion: Maximizing for domain invariance.
\newblock {\em arXiv preprint:1412.3474}, 2014.

\bibitem{wang2018visual}
Jindong Wang, Wenjie Feng, Yiqiang Chen, Han Yu, Meiyu Huang, and Philip~S Yu.
\newblock Visual domain adaptation with manifold embedded distribution
  alignment.
\newblock In {\em MM}, pages 402--410, 2018.

\bibitem{ganin2014unsupervised}
Yaroslav Ganin and Victor Lempitsky.
\newblock Unsupervised domain adaptation by backpropagation.
\newblock In {\em ICML}, 2015.

\bibitem{tzeng2017adversarial}
Eric Tzeng, Judy Hoffman, Kate Saenko, and Trevor Darrell.
\newblock Adversarial discriminative domain adaptation.
\newblock In {\em CVPR}, pages 7167--7176, 2017.

\bibitem{zhang2019bridging}
Yuchen Zhang, Tianle Liu, Mingsheng Long, and Michael~I Jordan.
\newblock Bridging theory and algorithm for domain adaptation.
\newblock In {\em ICML}, 2019.

\bibitem{courty2016optimal}
Nicolas Courty, R{\'e}mi Flamary, Devis Tuia, and Alain Rakotomamonjy.
\newblock Optimal transport for domain adaptation.
\newblock {\em IEEE TPAMI}, 39(9):1853--1865, 2016.

\bibitem{long2013transfer}
Mingsheng Long, Jianmin Wang, Guiguang Ding, Jiaguang Sun, and Philip~S Yu.
\newblock Transfer feature learning with joint distribution adaptation.
\newblock In {\em ICCV}, 2013.

\bibitem{wang2017balanced}
Jindong Wang, Yiqiang Chen, Shuji Hao, Wenjie Feng, and Zhiqi Shen.
\newblock Balanced distribution adaptation for transfer learning.
\newblock In {\em ICDM}, pages 1129--1134, 2017.

\bibitem{fang2019open}
Zhen Fang, Jie Lu, Feng Liu, Junyu Xuan, and Guangquan Zhang.
\newblock Open set domain adaptation: Theoretical bound and algorithm.
\newblock {\em IEEE TNNLS}, 2019.

\bibitem{yu2019transfer}
Chaohui Yu, Jindong Wang, Yiqiang Chen, and Meiyu Huang.
\newblock Transfer learning with dynamic adversarial adaptation network.
\newblock In {\em ICDM}, 2019.

\bibitem{johansson2019support}
Fredrik~D Johansson, David Sontag, and Rajesh Ranganath.
\newblock Support and invertibility in domain-invariant representations.
\newblock {\em arXiv preprint arXiv:1903.03448}, 2019.

\bibitem{gretton2012kernel}
Arthur Gretton, Karsten~M Borgwardt, Malte~J Rasch, Bernhard Sch{\"o}lkopf, and
  Alexander Smola.
\newblock A kernel two-sample test.
\newblock {\em JMLR}, 13(Mar):723--773, 2012.

\bibitem{ramdas2015decreasing}
Aaditya Ramdas, Sashank~Jakkam Reddi, Barnab{\'a}s P{\'o}czos, Aarti Singh, and
  Larry Wasserman.
\newblock On the decreasing power of kernel and distance based nonparametric
  hypothesis tests in high dimensions.
\newblock In {\em AAAI}, 2015.

\bibitem{hu2017unifying}
Zhiting Hu, Zichao Yang, Ruslan Salakhutdinov, and Eric~P Xing.
\newblock On unifying deep generative models.
\newblock {\em arXiv preprint arXiv:1706.00550}, 2017.

\bibitem{huszar2015not}
Ferenc Husz{\'a}r.
\newblock How (not) to train your generative model: Scheduled sampling,
  likelihood, adversary?
\newblock {\em arXiv preprint arXiv:1511.05101}, 2015.

\bibitem{theis2015note}
Lucas Theis, A{\"a}ron van~den Oord, and Matthias Bethge.
\newblock A note on the evaluation of generative models.
\newblock {\em arXiv preprint arXiv:1511.01844}, 2015.

\bibitem{wei2018transfer}
Ying Wei, Yu~Zhang, Junzhou Huang, and Qiang Yang.
\newblock Transfer learning via learning to transfer.
\newblock In {\em ICML}, pages 5085--5094, 2018.

\bibitem{csaji2001approximation}
Bal{\'a}zs~Csan{\'a}d Cs{\'a}ji.
\newblock Approximation with artificial neural networks.
\newblock {\em Faculty of Sciences, Etvs Lornd University, Hungary}, 24:48,
  2001.

\bibitem{pan2010survey}
Sinno~Jialin Pan, Qiang Yang, et~al.
\newblock A survey on transfer learning.
\newblock {\em IEEE TKDE}, 22(10):1345--1359, 2010.

\bibitem{zellinger2017central}
Werner Zellinger, Thomas Grubinger, Edwin Lughofer, Thomas Natschl{\"a}ger, and
  Susanne Saminger-Platz.
\newblock Central moment discrepancy (cmd) for domain-invariant representation
  learning.
\newblock In {\em ICLR}, 2017.

\bibitem{sun2016deep}
Baochen Sun and Kate Saenko.
\newblock Deep coral: Correlation alignment for deep domain adaptation.
\newblock In {\em ECCV}, pages 443--450, 2016.

\bibitem{courty2014domain}
Nicolas Courty, R{\'e}mi Flamary, and Devis Tuia.
\newblock Domain adaptation with regularized optimal transport.
\newblock In {\em ECML PKDD}, pages 274--289. Springer, 2014.

\bibitem{courty2017joint}
Nicolas Courty, R{\'e}mi Flamary, Amaury Habrard, and Alain Rakotomamonjy.
\newblock Joint distribution optimal transportation for domain adaptation.
\newblock In {\em NIPS}, pages 3730--3739, 2017.

\bibitem{bhushan2018deepjdot}
Bharath Bhushan~Damodaran, Benjamin Kellenberger, R{\'e}mi Flamary, Devis Tuia,
  and Nicolas Courty.
\newblock Deepjdot: Deep joint distribution optimal transport for unsupervised
  domain adaptation.
\newblock In {\em ECCV}, pages 447--463, 2018.

\bibitem{zhang2019optimal}
Zhen Zhang, Mianzhi Wang, and Arye Nehorai.
\newblock Optimal transport in reproducing kernel hilbert spaces: Theory and
  applications.
\newblock {\em IEEE TPAMI}, 2019.

\bibitem{gong2012geodesic}
Boqing Gong, Yuan Shi, Fei Sha, and Kristen Grauman.
\newblock Geodesic flow kernel for unsupervised domain adaptation.
\newblock In {\em CVPR}, pages 2066--2073. IEEE, 2012.

\bibitem{sun2015subspace}
Baochen Sun and Kate Saenko.
\newblock Subspace distribution alignment for unsupervised domain adaptation.
\newblock In {\em BMVC}, pages 24--1, 2015.

\bibitem{sun2016return}
Baochen Sun, Jiashi Feng, and Kate Saenko.
\newblock Return of frustratingly easy domain adaptation.
\newblock In {\em AAAI}, volume~6, page~8, 2016.

\bibitem{goodfellow2014generative}
Ian Goodfellow, Jean Pouget-Abadie, Mehdi Mirza, Bing Xu, David Warde-Farley,
  Sherjil Ozair, Aaron Courville, and Yoshua Bengio.
\newblock Generative adversarial nets.
\newblock In {\em NIPS}, pages 2672--2680, 2014.

\bibitem{luo2018transfer}
Yong Luo, Yonggang Wen, Ling-Yu Duan, and Dacheng Tao.
\newblock Transfer metric learning: Algorithms, applications and outlooks.
\newblock {\em arXiv preprint arXiv:1810.03944}, 2018.

\bibitem{ioffe2015batch}
Sergey Ioffe and Christian Szegedy.
\newblock Batch normalization: Accelerating deep network training by reducing
  internal covariate shift.
\newblock In {\em ICML}, 2015.

\bibitem{cariucci2017autodial}
Fabio~Maria Cariucci, Lorenzo Porzi, Barbara Caputo, Elisa Ricci, and
  Samuel~Rota Bulo.
\newblock Autodial: Automatic domain alignment layers.
\newblock In {\em ICCV}, pages 5077--5085. IEEE, 2017.

\bibitem{li2018adaptive}
Yanghao Li, Naiyan Wang, Jianping Shi, Xiaodi Hou, and Jiaying Liu.
\newblock Adaptive batch normalization for practical domain adaptation.
\newblock {\em Pattern Recognition}, 80:109--117, 2018.

\bibitem{wang2019transferable}
Ximei Wang, Ying Jin, Mingsheng Long, Jianmin Wang, and Michael~I Jordan.
\newblock Transferable normalization: Towards improving transferability of deep
  neural networks.
\newblock In {\em NeurIPS}, pages 1951--1961, 2019.

\bibitem{bousmalis2017unsupervised}
Konstantinos Bousmalis, Nathan Silberman, David Dohan, Dumitru Erhan, and Dilip
  Krishnan.
\newblock Unsupervised pixel-level domain adaptation with generative
  adversarial networks.
\newblock In {\em CVPR}, pages 3722--3731, 2017.

\bibitem{sugiyama2015introduction}
Masashi Sugiyama.
\newblock {\em Introduction to statistical machine learning}.
\newblock Morgan Kaufmann, 2015.

\bibitem{jing2020self}
Longlong Jing and Yingli Tian.
\newblock Self-supervised visual feature learning with deep neural networks: A
  survey.
\newblock {\em IEEE Transactions on Pattern Analysis and Machine Intelligence},
  2020.

\bibitem{long2018conditional}
Mingsheng Long, Zhangjie Cao, Jianmin Wang, and Michael~I Jordan.
\newblock Conditional adversarial domain adaptation.
\newblock In {\em NeurIPS}, pages 1640--1650, 2018.

\bibitem{imageclef}
The imageclef-da challenge 2014.
\newblock https://www.imageclef.org/2014.

\bibitem{venkateswara2017deep}
Hemanth Venkateswara, Jose Eusebio, Shayok Chakraborty, and Sethuraman
  Panchanathan.
\newblock Deep hashing network for unsupervised domain adaptation.
\newblock In {\em CVPR}, pages 5018--5027, 2017.

\bibitem{visda2017}
Xingchao Peng, Ben Usman, Neela Kaushik, Judy Hoffman, Dequan Wang, and Kate
  Saenko.
\newblock Visda: The visual domain adaptation challenge, 2017.

\bibitem{saenko2010adapting}
Kate Saenko, Brian Kulis, Mario Fritz, and Trevor Darrell.
\newblock Adapting visual category models to new domains.
\newblock In {\em ECCV}, pages 213--226. Springer, 2010.

\bibitem{he2016deep}
Kaiming He, Xiangyu Zhang, Shaoqing Ren, and Jian Sun.
\newblock Deep residual learning for image recognition.
\newblock In {\em CVPR}, pages 770--778, 2016.

\bibitem{long2016deep}
Mingsheng Long, Han Zhu, Jianmin Wang, and Michael~I Jordan.
\newblock Deep transfer learning with joint adaptation networks.
\newblock In {\em ICML}, 2017.

\bibitem{pei2018multi}
Zhongyi Pei, Zhangjie Cao, Mingsheng Long, and Jianmin Wang.
\newblock Multi-adversarial domain adaptation.
\newblock In {\em AAAI}, 2018.

\bibitem{zhang2018collaborative}
Weichen Zhang, Wanli Ouyang, Wen Li, and Dong Xu.
\newblock Collaborative and adversarial network for unsupervised domain
  adaptation.
\newblock In {\em CVPR}, pages 3801--3809, 2018.

\bibitem{you2019towards}
Kaichao You, Ximei Wang, Mingsheng Long, and Michael Jordan.
\newblock Towards accurate model selection in deep unsupervised domain
  adaptation.
\newblock In {\em ICML}, pages 7124--7133, 2019.

\bibitem{saito2018maximum}
Kuniaki Saito, Kohei Watanabe, Yoshitaka Ushiku, and Tatsuya Harada.
\newblock Maximum classifier discrepancy for unsupervised domain adaptation.
\newblock In {\em CVPR}, pages 3723--3732, 2018.

\bibitem{sankaranarayanan2018generate}
Swami Sankaranarayanan, Yogesh Balaji, Carlos~D Castillo, and Rama Chellappa.
\newblock Generate to adapt: Aligning domains using generative adversarial
  networks.
\newblock In {\em CVPR}, pages 8503--8512, 2018.

\bibitem{zhang2020covid}
Jianpeng Zhang, Yutong Xie, Yi~Li, Chunhua Shen, and Yong Xia.
\newblock Covid-19 screening on chest x-ray images using deep learning based
  anomaly detection.
\newblock {\em arXiv preprint arXiv:2003.12338}, 2020.

\bibitem{zhang2020covidda}
Yifan Zhang, Shuaicheng Niu, Zhen Qiu, Ying Wei, Peilin Zhao, Jianhua Yao,
  Junzhou Huang, Qingyao Wu, and Mingkui Tan.
\newblock Covid-da: Deep domain adaptation from typical pneumonia to covid-19.
\newblock {\em arXiv preprint arXiv:2005.01577}, 2020.

\bibitem{li2015generative}
Yujia Li, Kevin Swersky, and Rich Zemel.
\newblock Generative moment matching networks.
\newblock In {\em ICML}, 2015.

\bibitem{pham2020meta}
Hieu Pham, Qizhe Xie, Zihang Dai, and Quoc~V Le.
\newblock Meta pseudo labels.
\newblock {\em arXiv preprint arXiv:2003.10580}, 2020.

\bibitem{jiang2017mentornet}
Lu~Jiang, Zhengyuan Zhou, Thomas Leung, Li-Jia Li, and Li~Fei-Fei.
\newblock Mentornet: Learning data-driven curriculum for very deep neural
  networks on corrupted labels.
\newblock In {\em ICML}, 2018.

\bibitem{shu2019meta}
Jun Shu, Qi~Xie, Lixuan Yi, Qian Zhao, Sanping Zhou, Zongben Xu, and Deyu Meng.
\newblock Meta-weight-net: Learning an explicit mapping for sample weighting.
\newblock In {\em NeurIPS}, 2019.

\bibitem{ganin2016domain}
Yaroslav Ganin, Evgeniya Ustinova, Hana Ajakan, Pascal Germain, Hugo
  Larochelle, Fran{\c{c}}ois Laviolette, Mario Marchand, and Victor Lempitsky.
\newblock Domain-adversarial training of neural networks.
\newblock {\em JMLR}, 17(1):2096--2030, 2016.

\bibitem{long2016unsupervised}
Mingsheng Long, Han Zhu, Jianmin Wang, and Michael~I Jordan.
\newblock Unsupervised domain adaptation with residual transfer networks.
\newblock In {\em NIPS}, pages 136--144, 2016.

\bibitem{cai2019learning}
Ruichu Cai, Zijian Li, Pengfei Wei, Jie Qiao, Kun Zhang, and Zhifeng Hao.
\newblock Learning disentangled semantic representation for domain adaptation.
\newblock In {\em IJCAI}, 2019.

\bibitem{chen2020adversarial}
Minghao Chen, Shuai Zhao, Haifeng Liu, and Deng Cai.
\newblock Adversarial-learned loss for domain adaptation.
\newblock In {\em AAAI}, 2020.

\bibitem{lecun1998gradient}
Yann LeCun, L{\'e}on Bottou, Yoshua Bengio, and Patrick Haffner.
\newblock Gradient-based learning applied to document recognition.
\newblock {\em Proceedings of the IEEE}, 86(11):2278--2324, 1998.

\bibitem{snoek2012practical}
Jasper Snoek, Hugo Larochelle, and Ryan~P Adams.
\newblock Practical bayesian optimization of machine learning algorithms.
\newblock In {\em NIPS}, pages 2951--2959, 2012.

\end{thebibliography}

%\documentclass{article}
%
%% if you need to pass options to natbib, use, e.g.:
%%     \PassOptionsToPackage{numbers, compress}{natbib}
%     \PassOptionsToPackage{numbers, sort, compress}{natbib}
%% before loading neurips_2020
%
%% ready for submission
%\usepackage{neurips_2020}
%
%% to compile a preprint version, e.g., for submission to arXiv, add add the
%% [preprint] option:
%    %  \usepackage[preprint]{neurips_2020}
%
%% to compile a camera-ready version, add the [final] option, e.g.:
%%     \usepackage[final]{neurips_2020}
%
%% to avoid loading the natbib package, add option nonatbib:
%    %  \usepackage[nonatbib]{neurips_2020}
%
%\usepackage[utf8]{inputenc} % allow utf-8 input
%\usepackage[T1]{fontenc}    % use 8-bit T1 fonts
%\usepackage{hyperref}       % hyperlinks
%\usepackage{url}            % simple URL typesetting
%\usepackage{booktabs}       % professional-quality tables
%\usepackage{amsfonts}       % blackboard math symbols
%\usepackage{nicefrac}       % compact symbols for 1/2, etc.
%\usepackage{microtype}      % microtypography
%
%\usepackage{graphicx}
%\usepackage{subfigure}
%\usepackage{amsmath}
%\usepackage{amssymb}
%\usepackage{multirow}
%\usepackage{algorithm}
%\usepackage{algorithmic}

\appendix
\section*{Supplementary: Learning to Match Distributions for Domain Adaptation}

%\begin{document}
\maketitle
\setcounter{figure}{0}
\setcounter{table}{0}
\setcounter{equation}{0}
\section{Learning algorithm for L2M}

We also put the framework and key learning steps of L2M here for better illustration. The complete learning procedure of L2M is listed in Algorithm~\ref{alg:algorithm}.

\begin{figure}[htbp!]
	\centering\includegraphics[width=\textwidth]{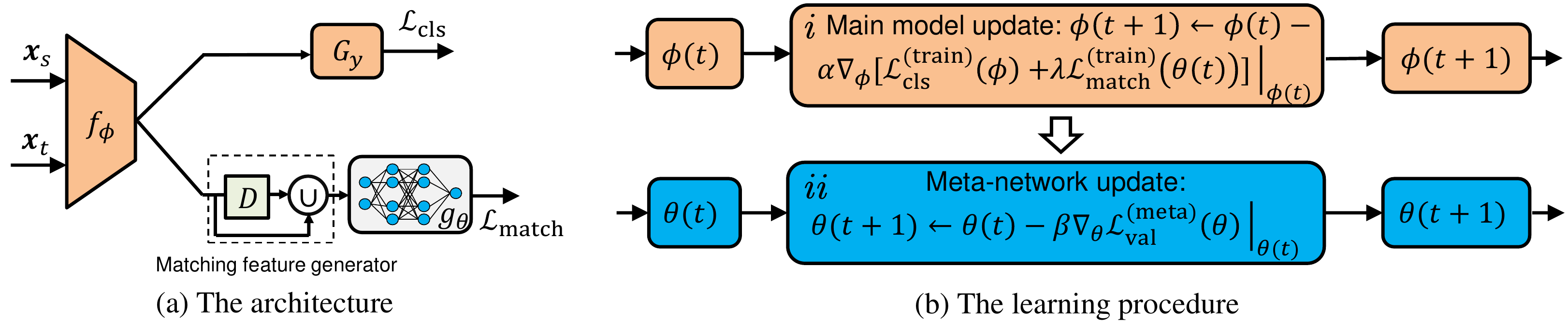}
	\vspace{-.1in}
	\caption{The framework and computing flow of the proposed L2M approach.}

\end{figure}

\begin{algorithm}[htb]
\caption{Learning algorithm of L2M}
\label{alg:algorithm}
\textbf{Input}: Source domain $\mathcal{D}_{s} = \{(\mathbf{x}^{s}_{i},y^{s}_{i})\}^{n_s}_{i=1}$, target domain $\mathcal{D}_{t} = \{\mathbf{x}^{t}_{j}\}^{n_t}_{j=1}$, learning rate $\alpha, \beta$, max epochs $T$.
		        
\textbf{Output}: $\{\phi^\star, \theta^\star\}$.
\begin{algorithmic}[1] %[1] enables line numbers
\STATE Initialize $\phi(0)$ and $\theta(0)$.
\WHILE{epoch $t < T$}
    \STATE Build an assist model with its parameter inherited from the main model $\phi(t)$.
    \STATE Sample a mini-batch data $B_s, B_t$ from both the source and target domain.
    \STATE Update $\phi$ by step $\romannumeral1$ in Fig.~\ref{fig-framework}(b). The loss consists of $\mathcal{L}_{\mathrm{cls}}$ and $\mathcal{L}_{\mathrm{match}}$, we only update the assist model $\phi$ and the meta-network $\theta$ only be updated in step $\romannumeral2$.
    \STATE Select the data with the highest prediction confidence from $\mathcal{D}_t$ to construct meta-data $\mathcal{D}_{\mathrm{meta}}$.
    \STATE Update the meta-network $\theta$ by step $\romannumeral2$ in Fig.~\ref{fig-framework}(b). 
\ENDWHILE
    \STATE \textbf{return} $\{\phi^\star, \theta^\star\}$
\end{algorithmic}
\end{algorithm}

It is worth noting that this optimization is general and can be naturally used in image generation tasks. Hence, we also use the same optimization step in the MNIST digits generation experiments by injecting this process directly to the GMMN~\cite{li2015generative} models. Therefore, L2M is a general and flexible framework that can work for most cross-domain distribution matching tasks.

\subsection{Matching features}
\paragraph{Task-independent matching features.}
It is natural to use the extracted feature embedding by $f_\phi$ as one kind of task-independent features, which is denoted as $\mathbf{F}_{\operatorname{emb}} \in \mathbb{R}^{d}$, where $d$ is the number of neurons in this layer. For classification tasks, another kind of features is the network logit: $\mathbf{F}_{\operatorname{logit}} \in \mathbb{R}^{C}$, which is the activation of the last FC layer before softmax. Note that in fact, $\mathbf{F}_{\operatorname{logit}}$ should be computed by $G_y$. For symbolic brievity we also draw it in the same way as $\mathbf{F}_{\operatorname{emb}}$ in Fig.~\ref{fig-framework}(a). Denote $q$ the function of last FC layer, then they can be computed as:
\begin{equation}
\label{eq-emb}
	\mathbf{F}_{\operatorname{emb}}=[f_\phi(\mathbf{x}_i). f_\phi(\mathbf{x}_j)].
\end{equation}
\begin{equation}
	\mathbf{F}_{\operatorname{logit}}=[q(f_\phi(\mathbf{x}_i)). q(f_\phi(\mathbf{x}_j))].
\end{equation}

\paragraph{Human-designed matching features.}
We adopt two popular distances as human-designed matching features: explicit distribution matching distance using MMD ($\mathbf{F}_{\operatorname{mmd}} \in \mathbb{R}$), and implicit distribution matching distance using adversarial nets ($\mathbf{F}_{\operatorname{adv}} \in \mathbb{R}$). Their basic idea is to approximate the joint distributions using marginal or conditional distributions. A recent work MEDA~\cite{wang2018visual} showed that matching both conditional and marginal distributions can be useful. Therefore, we denote $d_\mathrm{m}, d_\mathrm{c}$ the marginal and conditional distances (losses) respectively. Then, these features can be computed as:
\begin{equation}
\label{eq-mmd}
    \mathbf{F}_{\operatorname{mmd}} = [d_\mathrm{m}(f_\phi(\mathbf{x}_i),f_\phi(\mathbf{x}_j)), d_\mathrm{c}(f_\phi(\mathbf{x}_i),f_\phi(\mathbf{x}_j))].
\end{equation}
\begin{equation}
\label{eq-mmd}
    \mathbf{F}_{\operatorname{adv}} = [\mathbb{D}_\mathrm{m}(f_\phi(\mathbf{x}_i),f_\phi(\mathbf{x}_j)), \mathbb{D}_\mathrm{c}(f_\phi(\mathbf{x}_i),f_\phi(\mathbf{x}_j))].
\end{equation}

For explicit distribution matching using MMD~\cite{gretton2012kernel}, the marginal and conditional distances can be computed as:
\begin{equation}
    \begin{split}
    &d_{\mathrm{m}}=
\begin{Vmatrix}
\mathbb{E}_{\mathbf{x} \sim B_s} \phi(\mathbf{x}) - \mathbb{E}_{\mathbf{x} \sim B_t} \phi(\mathbf{x})
\end{Vmatrix}^2_{\mathcal{H}_k},\\
&d_{\mathrm{c}}=\mathbb{E}_{c\sim \mathcal{C}}
\begin{Vmatrix}
\mathbb{E}_{\mathbf{x} \sim B^{(c)}_s} \phi(\mathbf{x}) - \mathbb{E}_{\mathbf{x} \sim B^{(c)}_t} \phi(\mathbf{x})
\end{Vmatrix}^2_{\mathcal{H}_k},
    \end{split}
\end{equation}
where $\mathcal{H}_k$ is the Reproducing Kernel Hilbert Space (RKHS) induced by kernel $k$, $B^{(c)}$ denotes samples belonging to class $c$, and $\phi(\cdot)$ some feature mapping function. 

For implicit distribution matching using GAN~\cite{goodfellow2014generative}, the main idea is to design a domain discriminator $G_d$ to identify which domain the samples belong to. We train $f_\phi$ and $G_y$ to confuse $G_d$, and eventually $G_d$ gets confused and fails to discriminate the domains. In this situation, the marginal and conditional adversarial distances can be respectively computed as:
\begin{equation}
    \begin{split}
    &\mathbb{D}_{\mathrm{m}}=
\mathbb{E}_{\mathbf{x} \sim B_{s}\cup B_{t}} \ell_{d}(G_{d}(f_{\phi}(\mathbf{x})),d),\\
&\mathbb{D}_{\mathrm{c}}=\mathbb{E}_{c\sim \mathcal{C}} \mathbb{E}_{\mathbf{x} \sim B_{s}\cup B_{t}}
 \ell^{(c)}_{d}(G^{(c)}_{d}(\hat{y}^{(c)} f_{\phi} (\mathbf{x})),d),
    \end{split}
\end{equation}
where $\ell_{d}$ is the cross-entropy loss for domain classification, and $d$ is the domain label (0 or 1) of the input sample $\mathbf{x}$.
$G^{(c)}_{d}$ and $\ell^{(c)}_{d}$ are the conditional domain discriminator and its cross-entropy loss associated with class $c$, respectively. $\hat{y}^{(c)}$ is the predicted label over the class $c$ of the input sample $\mathbf{x}$.

Note that the target domain $\mathcal{D}_t$ has no labels, making it difficult to compute the conditional distance $d_c$. We apply prediction to $\mathcal{D}_t$ using the classifier $G_y$ trained on $\mathcal{D}_s$ to obtain soft labels, which will be iteratively refined. Clearly, MMD and adversarial distance are only two options for predefined distance and others can be used. In specific problems, more task-dependent features can be used. This makes L2M a general and flexible framework.

\subsection{Convergence analysis and theoretical insights}
In addition to empirically analyzing the convergence of L2M, we provide some theoretical analysis. The convergence of L2M depends on two items: the classification loss $\mathcal{L}_{\operatorname{cls}}$ on the training data, and the distribution matching loss $\mathcal{L}_{\operatorname{match}}$ on the meta-data. The convergence of $\mathcal{L}_{\operatorname{cls}}$ is well ensured since it is a standard cross-entropy loss in deep neural networks. The convergence of $\mathcal{L}_{\operatorname{match}}$ depends on two factors: the construction of meta-data and the loss itself. We adopt an iterative way to construct the meta-data by using the pseudo labels provided by the trained network. According to several recent works~\cite{wang2018visual,long2018conditional,zhang2019bridging}, the convergence of such an iterative pseudo-label can be ensured, i.e., the pseudo labels will be more accurate, providing a strong support to the construction of the meta-data. On the other hand, the convergence of $\mathcal{L}_{\operatorname{match}}$ can also be ensured as long as the meta-network $g_\theta$ is differential (in our work, it is differential) by following~\cite{jiang2017mentornet,shu2019meta}. Therefore, the convergence of $\mathcal{L}_{\operatorname{match}}$ can be ensured.

In the view of domain adaptation theory, L2M is designed by following the DA theory according to~\cite{ben2007analysis} that the risk on the target domain is bounded by the following theorem:

\textbf{Theorem 1} \textit{Let $h \in \mathcal{H}$ be a hypothesis, $\epsilon_s(h)$ and $\epsilon_t(h)$ be the expected risks on the source and target domain, respectively, then}
\begin{equation}
    \label{eq-theorem}
    \epsilon_{t}(h) \leqslant \epsilon_{s}(h)+d_{\mathcal{H}}(p, q)+C_{0},
\end{equation}
where $C_0$ is a constant for the complexity of hypothesis and plus the risk of an ideal hypothesis for both domains. $d_{\mathcal{H}}(p, q)$ refers to the distribution divergence between domains. As can be seen, L2M is directly minimizing the distribution distance (distribution matching loss) $\mathcal{L}_{\operatorname{match}}$, which is in consistence with the above theorem.

\subsection{Remarks}

\paragraph{L2M vs. AutoML.}
L2M shares the same goal with AutoML: both are trying to reduce the human intervention in a machine learning process. However, AutoML focuses more on ``auto'' while L2M can be seen as a combination of deep features and human-designed features. Moreover, AutoML focuses on architecture design, hyperparameter search, and channel pruning, which are different from L2M. The main goal of L2M is to learn a good and automatic distribution matching between domains. From this point of view, L2M can also be seen as an ``automated'' DA method. Future works may lay emphasis on domain adaptation architecture design, which is more like automl.

\paragraph{L2M is not yet another ``SOTA'' and is not intending replacing other methods.}
The results in this paper demonstrates that the performance of L2M outperforms several SOTA methods. However, our goal is not to develop yet another SOTA to the community, but to introduce another kind of DA algorithm that can be easily applied to real applications without specific concentration on the loss function design and distribution matching module. Therefore, for a new application, both L2M and other existing SOTA methods are applicable. The advantage of using L2M is that it requires less human intervention of algorithm selection, while a simple embedding matching feature can achieve a competitive performance. If you need better results, you still need to have a deep domain knowledge and integrate it in L2M with the embeddings or logits features. Therefore, L2M can be used to enhance other methods.

\section{Experimental Details}
\subsection{Datasets}

\paragraph{ImageCLEF-DA.} ImageCLEF-DA~\cite{imageclef} is a benchmark dataset for ImageCLEF 2014 domain adaptation challenge, and it is collected by selecting the 12 common categories shared by the following public datasets and each of them is considered as a domain: $Caltech-256$ (\textbf{C}), $ImageNet~ILSVRC~2012$ (\textbf{I}), $Pascal~VOC~2012$ (\textbf{P}). There are 50 images in each category and 600 images in each domain. We use all domain combinations and build 6 transfer tasks.

\paragraph{Office-Home.} Office-Home~\cite{venkateswara2017deep} consists of images from 4 different domains: $Artistic~images$ (\textbf{A}), $Clip~Art$ (\textbf{C}), $Product~images$ (\textbf{P}) and $Real-World~images$ (\textbf{R}). For each domain, the dataset contains images of 65 object categories collected in office and home settings. Similarly, we use all domain combinations and construct 12 transfer tasks.

\paragraph{VisDA-2017.} VisDA-2017~\cite{visda2017} is a simulation-to-real dataset with two extremely distinct domains: Synthetic renderings of 3D models and Real collected from photo-realistic or real-image datasets. With 280K images in 12 classes, the scale of VisDA-2017 brings challenges to domain adaptation.

\paragraph{Office-31.} Office-31 dataset~\cite{saenko2010adapting} is a standard and maybe the most popular benchmark for unsupervised domain adaptation. It consists of 4,110 images within 31 categories collected from everyday objects in an office environment. It consists of three domains: $Amazon$ (\textbf{A}), which contains images downloaded from \url{amazon.com}, $Webcam$ (\textbf{W}) and $DSLR$ (\textbf{D}), which contain images respectively taken by web camera and digital SLR camera under different settings. We evaluate all our methods across six transfer tasks on all three domains.

The statistics of these datasets are shown in Table~\ref{tb-dataset}. 

%% Information of the datasets
\begin{table}[t!]
	\caption{Public datasets description.}
	\vspace{-.1in}
	\label{tb-dataset}
	\centering
		\begin{tabular}{cccc}
			\toprule
			Dataset & \#Sample & \#Class & \#Domain \\ \hline
			Office-31 & 4,110 & 31 & A, W, D \\ 
			ImageCLEF-DA & 1,800 & 12 & C, I, P \\ 
			Office-Home & 15,588 & 65 & A, C, P, R \\ 
			VisDA-2017 & 280,000 & 12 & Synthetic, Real \\ \bottomrule
		\end{tabular}
	
	\vspace{-.1in}
\end{table}

\subsection{Implementation Details}

For different variants of L2M using different matching features, we report the dimension information of eight matching features of each dataset in Table~\ref{tb-data-di}. 

%% matching features of the datasets
\begin{table}[ht!]
	\caption{Dimension of matching features of the datasets.}
	\label{tb-data-di}
	\centering
	\resizebox{1.0\textwidth}{!}{
		\begin{tabular}{ccccccccc}
			\toprule
			Dataset & $\mathbf{F}_{\operatorname{emb}}$ & $\mathbf{F}_{\operatorname{logit}}$ & $\mathbf{F}_{\operatorname{mmd}}$ & $\mathbf{F}_{\operatorname{adv}}$ & $[\mathbf{F}_{\operatorname{emb}},\mathbf{F}_{\operatorname{mmd}}]$ & $[\mathbf{F}_{\operatorname{emb}},\mathbf{F}_{\operatorname{adv}}]$ & $[\mathbf{F}_{\operatorname{logit}},\mathbf{F}_{\operatorname{mmd}}]$ & $[\mathbf{F}_{\operatorname{logit}},\mathbf{F}_{\operatorname{adv}}]$  \\ \hline
			Office-31    & 2,048   &  31 & 2  &  2 & 2,050 & 2,050 & 33 & 33   \\ 
			ImageCLEF-DA & 2,048   &  12 & 2  &  2 & 2,050 & 2,050 & 14 & 14   \\ 
			Office-Home  & 2,048   &  65 & 2  &  2 & 2,050 & 2,050 & 67 & 67   \\ 
			VisDA-2017   & 2,048   &  12 & 2  &  2 & 2,050 & 2,050 & 14 & 14   \\ \bottomrule
		\end{tabular}
	}
	
\end{table}

All methods use the ImageNet-pretrained ResNet-50 as the backbone network. Results of the comparison methods are obtained from original papers. For L2M, we set max iterations to be 200000. The mini-batch SGD with nesterov momentum of 0.9 and batchsize 32 is used as the optimization strategy. The learning rate $\alpha$ of the meta-model and the overall model changes by following~\cite{ganin2014unsupervised}: $\alpha_{k} = \frac{\alpha}{(1+\gamma k)^{-\upsilon}}$, where $k$ is the training iteration linearly changing from 1 to max iterations, $\gamma=0.001$, $\alpha=0.004$, and decay rate $\upsilon=0.75$. The initial learning rate $\beta$ of the meta-network is 0.01 and will gradually decrease to 0.0001 during training. Meta-network $g_\theta$ uses a $d-1024-1024-1$ structure where $d$ is the dimension of input matching features, and more information of different matching features can be seen in Table~\ref{tb-data-di}. We follow the standard protocols for unsupervised domain adaptation~\cite{ganin2016domain}, we use classification accuracy on the target domain as the evaluation metric and target labels are only used for evaluation. The results are the average accuracy of 10 experiments by following the same protocol~\cite{zhang2019bridging,wang2018visual,tzeng2014deep,long2018conditional}. We use Pytorch to implement L2M and it is trained on a Linux machine with a 16GB P100 GPU.

\subsection{Results on Office-31 dataset}

Table~\ref{tb-supp-office31} reports the results on Office-31, which indicates that L2M outperforms all the recent DA methods in classification accuracy.

%% office-31
\begin{table}[t!]
    \caption{\upshape Accuracy~(\%) on Office-31 for UDA (ResNet-50).}
    \vspace{-.1in}
	\label{tb-supp-office31}
	\centering 
	\resizebox{.7\textwidth}{!}{
	\begin{tabular}{lccccccc}
		\toprule
		Method  & A$\rightarrow$W & A$\rightarrow$D & D$\rightarrow$W & D$\rightarrow$A & W$\rightarrow$D & W$\rightarrow$A & AVG \\ \hline
		ResNet~\cite{he2016deep}  & 68.4 & 68.9 & 96.7 & 62.5 & 99.3 & 60.7 & 76.1 \\
		DDC~\cite{tzeng2014deep}  & 75.6 & 76.5 & 96.0 & 62.2 & 98.2 & 61.5 & 78.3 \\
		DAN~\cite{long2015learning}  & 80.5 & 78.6 & 97.1 & 63.6 & 99.6 & 62.8 & 80.4 \\
 		D-CORAL~\cite{sun2016deep} & 77.0 & 81.5 & 97.1 & 65.9 & 99.6 & 64.3 & 80.9 \\
 		RTN~\cite{long2016unsupervised} & 84.5 & 77.5 & 96.8 & 66.2 & 99.4 & 64.8 & 81.6 \\
		DANN~\cite{ganin2014unsupervised} & 82.0 & 79.7 & 96.9 & 68.2 & 99.1 & 67.4 & 82.2 \\
		ADDA~\cite{tzeng2017adversarial} & 86.2 & 77.8 & 96.2 & 69.5 & 98.4 & 68.9 & 82.9 \\
		JAN~\cite{long2016deep} & 85.4 & 84.7 & 97.4 & 68.6 & 99.8 & 70.0 & 84.3 \\ 
		MADA~\cite{pei2018multi} & 90.0 & 87.8 & 97.4 & 70.3 & 99.6 & 66.4 & 85.2 \\
		MEDA~\cite{wang2018visual} & 86.0 & 86.3 & 97.1 & 72.1 & 99.2 & 73.2 & 85.7 \\
		CAN~\cite{zhang2018collaborative} & 92.5 & 90.1 & 98.8 & 72.1 & 100.0 & 69.9 & 87.2 \\
		DSR~\cite{cai2019learning}  & 93.1 & 92.4 & 98.7 & 73.5 & 99.8 & 73.9 & 88.6 \\
		CDAN~\cite{long2018conditional} & 94.1 & 92.9 & 98.6 & 71.0 & 100.0 & 69.3 & 87.7 \\
		ALDA~\cite{chen2020adversarial} & \textbf{95.6} & 94.0 & 97.7 & 72.2 & 100.0 & 72.5 & 88.7 \\
		MDD~\cite{zhang2019bridging} & 94.5 & 93.5 & 98.7 & 74.6 & 100.0 & 72.2 & 88.9 \\ \hline
%		CDAN+TransNorm~\cite{wang2019transferable} & 95.7 & 94.0 & 98.7 & 73.4 & 100.0 & 74.2 & 89.3 \\ \hline
		L2M   & 93.2 & \textbf{94.1} & \textbf{98.8} & \textbf{75.9} & \textbf{100.0} & \textbf{74.7} & \textbf{89.5} \\ 
		\bottomrule
	\end{tabular}}
	\vspace{-.1in}
\end{table}

\subsection{More ablation experiments of L2M}
We show more ablation experiments of L2M on Office-Home and ImageCLEF-DA in Table~\ref{tb-supp-officehome} and Table~\ref{tb-supp-clef}, respectively. We did not run ablation experiments on VisDA-17 since this dataset is rather larger and needs more computations. The ablation results on other datasets are enough for observing the patterns of L2M variants. Combining these results with that from the main paper, more insightful conclusions can be made. \textbf{(1)}~L2M achieves the best performance on multiple datasets, which indicates the efficiency of L2M. \textbf{(2)}~All the 4 variants of L2M can achieve competitive performance, implying the effectiveness of the meta-network on matching functions and L2M is able to fit a wide range of matching features. \textbf{(3)}~L2M~(emb+adv) outperforms the other 3 variants of L2M, which indicates L2M can learn more representative and transferable features by taking as input deep features and human-designed features.

Despite the performance on these public datasets, we want to emphasis that in real applications, L2M~(emb+adv) is perhaps not always the best matching features. Therefore, in order to achieve the best performance, users can try several combinations of matching features along with their own domain experience before finding the suitable features. Since the performance of most matching features are with a low variance, any matching feature can achieve competitive performance compared to existing methods.

%% OfficeHome
\begin{table}[t!]
    \caption{\upshape Accuracy~(\%) on Office-Home for UDA (ResNet-50).}
	\label{tb-supp-officehome}
	\centering
	\setlength{\tabcolsep}{0.5mm}{
	\resizebox{\textwidth}{!}{
	\begin{tabular}{lccccccccccccc}
		\toprule
		Method & A$\rightarrow$C  & A$\rightarrow$P  & A$\rightarrow$R  & C$\rightarrow$A  & C$\rightarrow$P  & C$\rightarrow$R  & P$\rightarrow$A  & P$\rightarrow$C  & P$\rightarrow$R  & R$\rightarrow$A  & R$\rightarrow$C  & R$\rightarrow$P  & AVG  \\ \hline
		L2M (emb)   & 55.8 & 73.9 & 77.8 & 61.4 & 72.9 & 73.1 & 62.5 & 54.5 & 79.3 & 71.1 & 60.2 & 83.2 & 69.1 \\
		L2M (logit)   & 56.2 & 70.2 & 71.7 & 57.7 & 68.1 & 71.1 & 60.7 & 57.1 & 79.4 & 71.6 & 61.5 & 83.6 & 67.4 \\
		L2M (logit+adv)   & 55.4 & 70.6 & 76.6 & 58.1 & 69.2 & 70.1 & 61.1 & 55.5 & 78.6 & 71.4 & 60.7 & 82.8 & 67.5 \\
		L2M (emb+adv)   & 56.1 & 75.0 & 79.3 & 62.8 & 73.3 & 73.7 & 63.8 & 54.1 & 80.6 & 71.8 & 60.2 & 83.5 & 69.5 \\ \bottomrule
	\end{tabular}}}
\end{table}

%% ImageCLEF
\begin{table}[t!]
    \caption{\upshape Accuracy~(\%) on ImageCLEF-DA for UDA (ResNet-50).}
	\label{tb-supp-clef}
	\centering 
	\setlength{\tabcolsep}{4.0mm}{
	\begin{tabular}{lccccccc}
		\toprule
		Method  & I$\rightarrow$P & P$\rightarrow$I & I$\rightarrow$C & C$\rightarrow$I & C$\rightarrow$P & P$\rightarrow$C & AVG \\ \hline
		L2M (emb)  & 78.0 & 90.5 & 96.2 & 91.4 & 78.3 & 94.1 & 88.1 \\
		L2M (logit)  & 76.2 & 88.0 & 96.8 & 89.5 & 76.8 & 94.5 & 87.0 \\
		L2M (logit+adv)   & 77.0 & 89.2 & 95.7 & 89.8 & 77.3 & 93.3 & 87.1 \\
		L2M (emb+adv)    & 78.7 & 91.0 & 97.0 & 92.0 & 79.7 & 96.0 & 89.1 \\ \bottomrule
	\end{tabular}
	}
\end{table}

\subsection{Feature visualization.}
\begin{figure}[ht!]
	\centering
	\subfigure[ResNet50]{
		\centering
		\includegraphics[scale=0.2]{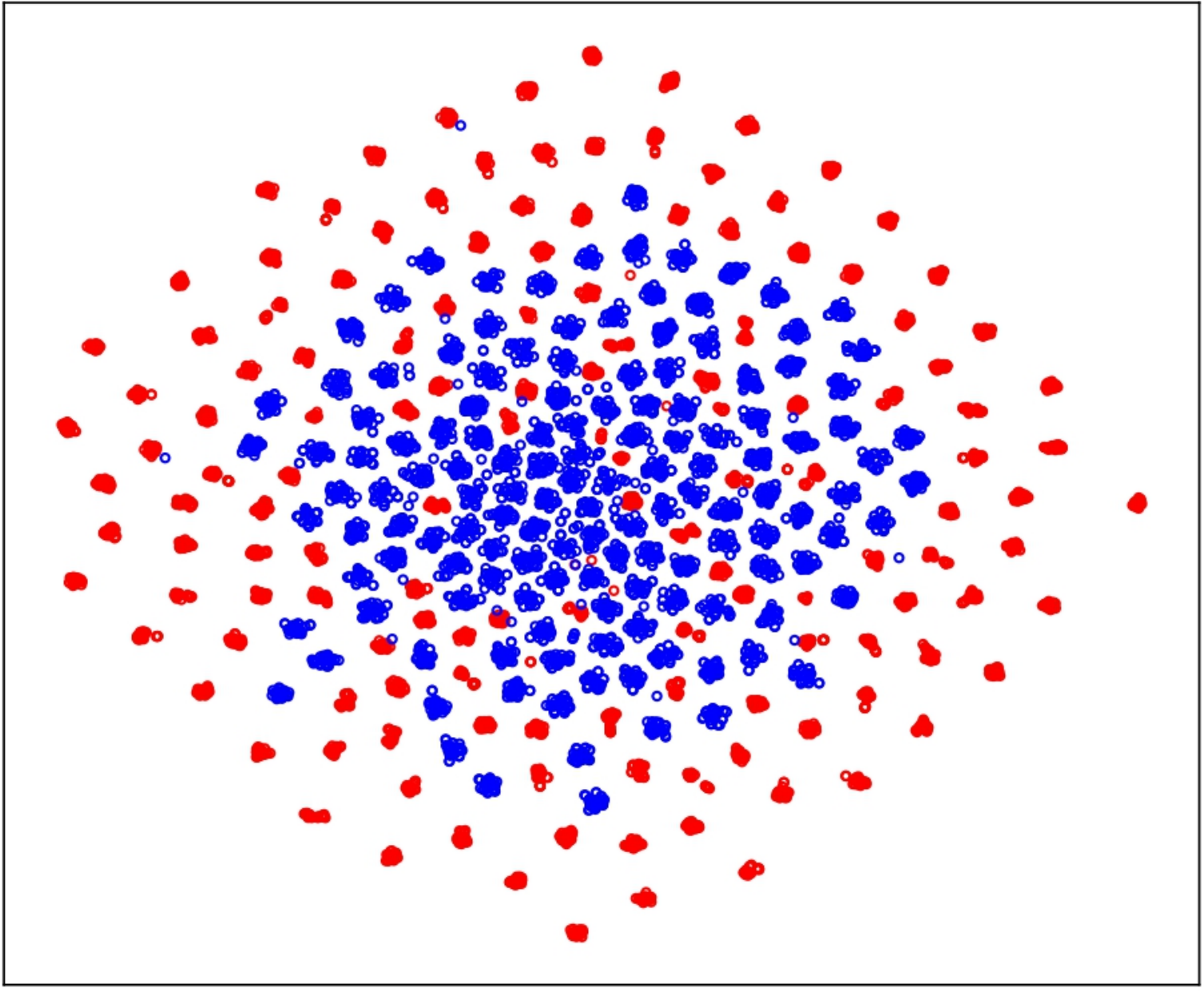}
		\label{fig-sub-resnet}}
	\subfigure[JAN]{
		\centering
		\includegraphics[scale=0.2]{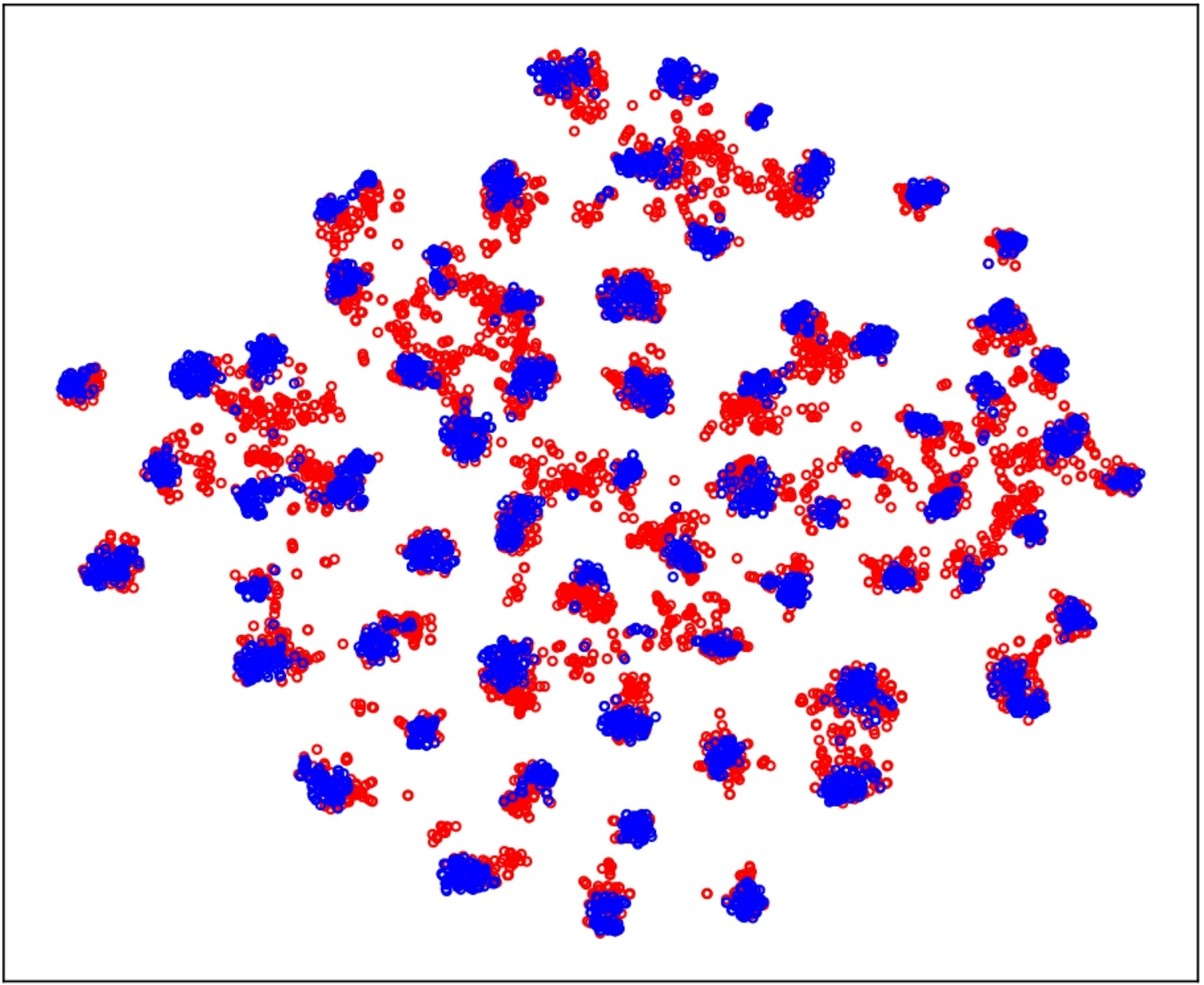}
		\label{fig-sub-jan}}
	\subfigure[MEDA]{
		\centering
		\includegraphics[scale=0.2]{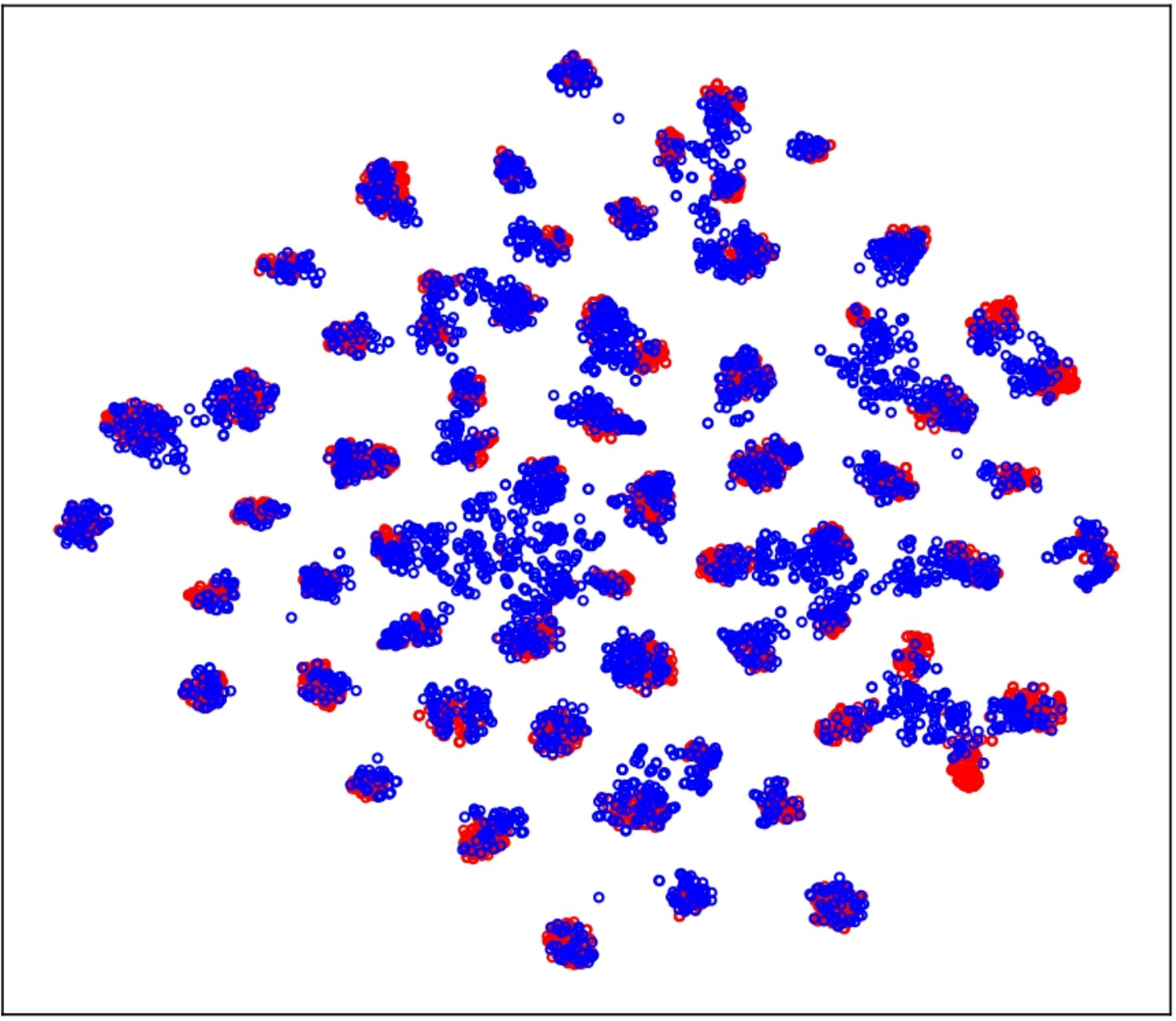}
		\label{fig-sub-daan}}
	\subfigure[L2M]{
		\centering
		\includegraphics[scale=0.2]{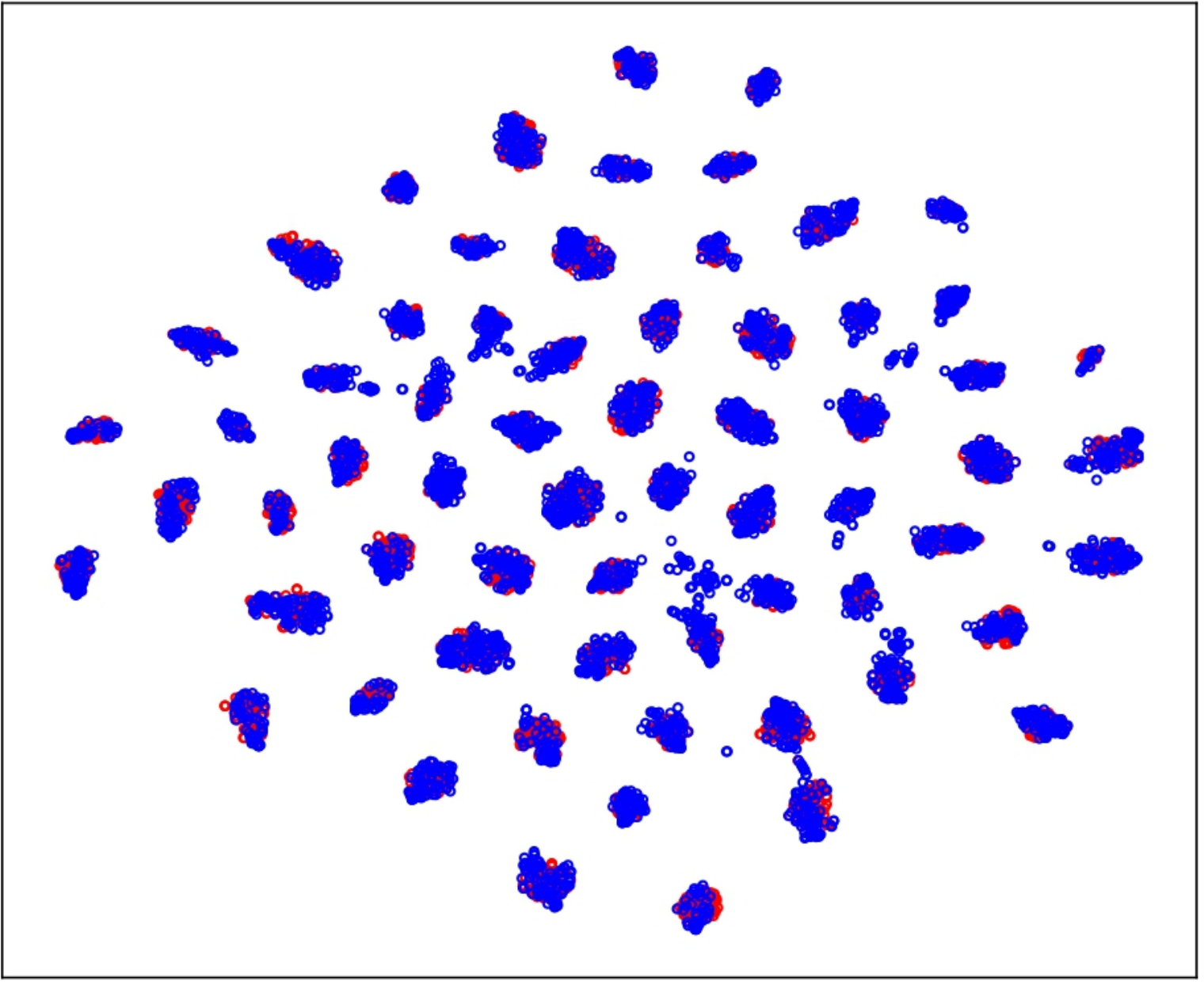}
		\label{fig-sub-metametch}}
	\vspace{-.1in}
	\caption{(Best viewed in color) The t-SNE visualization of network activation on task P$\rightarrow$R. Red circles are the source samples and blue circles are the target samples.}
	\vspace{-.1in}
	\label{fig-fvis}
\end{figure}

We visualize the network activation (before FC layer) on task P$\rightarrow$R using t-SNE in Fig.~\ref{fig-fvis}. ResNet-50 does not align the distributions. JAN aligns both marginal and conditional distributions with equal weights, while MEDA adaptively aligns these two distributions whose results are better. However, the source and target domains are not fully matched by MEDA. For L2M, both the cross-domain distributions and categories are aligned well, implying that L2M learns more discriminative features.

\section{Application to COVID-19 Chest X-ray Image Adaptation}

Other than benchmarking L2M on popular public datasets including Office-31, Office-Home, ImageCLEF-DA, and VisDA-17, we compare the performance of several DA methods including L2M in a real application. Different from public datasets, this application will prove the effectiveness of L2M and other DA methods in a real-world task, which is more appealing and inspiring.

We present more details for applying L2M to COVID-19 chest X-ray image adaptation tasks. COVID-19 is a specific type of pneumonia compared to the normal kind of pneumonia, and there is not too much COVID-19 data available, it becomes necessary and feasible to use the sufficient labeled pneumonia data to help classify the COVID-19 symptom. Therefore, this task is a binary classification task, where the source domain is the well-labeled pneumonia data to classify whether this patient is having pneumonia or not, and the target domain is the unlabeled COVID-19 data. Out task is to classify whether each of the the target domain samples is having a COVID-19 symptom or not.

This is a binary classification task, i.e., the normal category vs. pneumonia on the source domain, and the normal category vs. COVID-19 on the target domain. We also notice that this dataset is highly-imbalanced~(as shown in the next section). Therefore, for better illustrate the results, we adopt F1 score, Recall, and Precision as the evaluation metrics rather than classification accuracy. These metrics are better for imbalanced classification tasks. It also demonstrates our contribution that L2M can achieve robust preformance in imbalanced tasks compared to other DA methods.

\subsection{Dataset}

\begin{figure}[ht!]
    \centering
    \includegraphics[width=\textwidth]{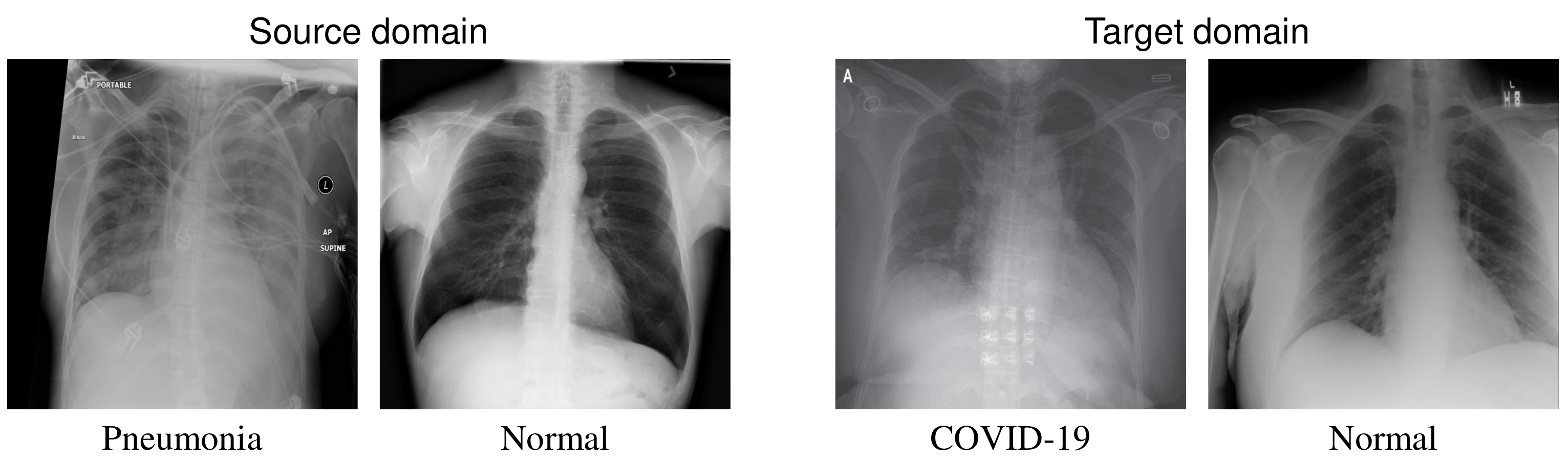}
    \caption{Samples from the source and target domain.}
    \label{fig-covid-data}
\end{figure}

Table~\ref{tb-covid-dataset} shows the description of the dataset\footnote{The dataset is available at https://github.com/qiuzhen8484/COVID-DA}. Note that in this task, we use some COVID-19 data as the validation set to better tune the hyperparameters. In the source domain, there are two classes: normal and pneumonia, while there are normal and COVID-19 classes in the target and validation dataset. Fig.~\ref{fig-covid-data} shows some examples from the source and target domain. It is clear that data from two domains are very similar especially for pneumonia and COVID-19 classes. Therefore, it is feasible to perform domain adaptation or transfer learning between these two domains.

\begin{table}[ht!]
\centering
\caption{Dataset description of pneumonia and COVID-19 dataset}
\label{tb-covid-dataset}
\begin{tabular}{cccccc}
\toprule
Domain & Symptom & \#Normal & \#Pneumonia & \#COVID-19 & \#Total \\ \hline
Source & Pneumonia & 5,613 & 2,306 & 0 & 7,919 \\ 
Target & COVID-19 & 885 & 0 & 60 & 945 \\ 
Validation & COVID-19 & 254 & 0 & 25 & 279 \\ \bottomrule
\end{tabular}
\end{table}

\subsection{Baselines and experimental details}

We mainly compare the performance of L2M with three categories of methods: (1) Deep learning baselines, (2) Deep diagnostic methods, and (3) unsupervised domain adaptation methods. The deep learning baselines include three baselines:
\begin{itemize}
    \item Train on source: train a network on the source domain, and then apply the pretrained model on the target domain.
    \item Train on target: this is an ideal state since there are no labels for the target domain in our task. Therefore, we directly use several extra labeled COVID-19 data from the dataset (they are 30\% of the target domain data) and train a network on these data. Then, we can apply prediction on the target data.
    \item Fine-tuning: This is a combination of the above two baselines. Firstly, we train a network on the source domain. Then, we fine-tune the pretrained model on the extra labeled target domain data. Finally, we apply prediction on the target data.
\end{itemize}
The deep diagnostic method is DLAD~\cite{zhang2020covid}.

The unsupervised DA methods are DANN~\cite{ganin2014unsupervised}, MCD~\cite{saito2018maximum}, and CDAN+TransNorm~\cite{wang2019transferable}. All methods are using ResNet-18 as the backbone network following~\cite{zhang2020covidda}. The results of these methods are obtained from COVID-DA~\cite{zhang2020covidda} to ensure a fair comparison. Note that we did not compare COVID-DA since this method is a semi-supervised method that explicitly uses the labeled data on the target domain. Therefore, we only use the report of unsupervised methods and train L2M with the same experimental settings.

\subsection{Analysis of the results}

The results are shown in Table~\ref{tb-covid}. Here we use the 95\% confidence interval, where the corresponding value of $z$ is 1.96. The computed confidence interval $r$ is around 1.3\%.

This table is the same as the main paper but with more analysis of the results. From the results, we see that L2M outperforms all comparison methods in terms of F1 score and Recall. In Precision, the performance of Training on labeled target data achieves the best results, which is reasonable since this approach trains on the labeled target domain data and is expected to achieve the best precision. The UDA methods, namely DANN, MCD, and CDAN+TransNorm can sometimes achieve worse results than baselines, indicating that the different distribution distance of pneumonia and COVID-19 data are not that easy to compute by adversarial distance (DANN and MCD are using adversarial distances) or statistical alignment (TransNorm uses a source-target normalization technique) since these methods are built with their own priors and biases. In this situation, it is necessary to perform domain adaptation in a data-driven way stepping back from these predefined distances. Therefore, L2M can be useful in real-world applications. It also demonstrates our contribution that L2M can achieve robust preformance in imbalanced tasks compared to other DA methods.

%% Covid
\begin{table}[ht!]
	\caption{\upshape Results on COVID-19 X-ray adaptation (normal pneumonia $\rightarrow$ COVID-19, ResNet-18). Here we use the 95\% confidence interval, where the corresponding value of $z$ is 1.96. The computed confidence interval $r$ is around 1.3\%.}
	\label{tb-covid}
	\centering
	\setlength{\tabcolsep}{5.0mm}{
		\resizebox{0.7\textwidth}{!}{
		\begin{tabular}{lccc}
		\toprule
		Method & Precision (\%) & Recall (\%) & F1 (\%) \\ \hline
		Train on source & 63.5 & 66.7 & 65.0 \\
		Train on target \textit{(ideal state)} & \textbf{91.7} & 55.0 & 68.8 \\
		Fine-tuning & 56.3 & 75.0 & 64.3 \\
		DLAD~\cite{zhang2020covid} & 62.0 & 73.3 & 67.2 \\
		DANN~\cite{ganin2014unsupervised} & 61.4 & 71.7 & 66.2 \\
		MCD~\cite{saito2018maximum} & 63.2 & 60.0 & 61.5 \\
		CDAN+TransNorm~\cite{wang2019transferable} & 85.0 & 39.2 & 63.7 \\ \hline
		L2M & 70.1 & \textbf{78.3} & \textbf{74.0} \\
		\bottomrule
        \end{tabular}
	}}
	\vspace{-.1in}
\end{table}

On the other hand, we also notice that the performance of fine-tuning is worse than training on source and target, which is probably due to the distribution gap between the source and target domains. In a nutshell, among baselines and DA methods other than L2M, training on target achieves the best performance, indicating the importance of labeled data. We can also see that in COVID-DA~\cite{zhang2020covidda}, authors used semi-supervised settings to improve the F1, Precision, and Recall score to over 90\%, which clearly shows optimistic performance. Therefore, L2M and other methods can also be applied to semi-supervised DA tasks by adopting several labeled target samples. This is left for future work since this work mainly focuses on unsupervised DA. 

\subsection{Ablation study}

We show the ablation study of L2M on this COVID-19 data in Table~\ref{tb-ablation-covid}. It is shown that by combining different matching features, L2M can generally achieve better performance than the comparison methods. 

\begin{table}[ht!]
\centering
\caption{Ablation experiments of L2M on COVID-19 experiment. Here we use the 95\% confidence interval, where the corresponding value of $z$ is 1.96. The computed confidence interval $r$ is around 1.3\%.}
\label{tb-ablation-covid}
\begin{tabular}{cccc}
\toprule
L2M variant & F1 & Recall & Precision \\ \hline
L2M (emb) & 73.2 & 75.0 & 71.4 \\ 
L2M (logit) & 68.3 & 70.0 & 66.7 \\ 
L2M (emb+mmd) & 69.1 & 78.3 & 61.8 \\ 
L2M (logit+mmd) & 74.0 & 78.3 & 72.3 \\ 
% L2M (emb(adv)) & 66.67 & 68.30 & 65.00 \\ 
% L2M (logit(adv)) & 67.70 & 70.00 & 65.62 \\ 
L2M (emb+adv) & 68.9 & 65.4 & 78.5 \\ 
L2M (logit+adv) & 69.4 & 71.7 & 67.2 \\ 
L2M (mmd) & 71.2 & 70.0 & 72.4 \\ 
L2M (adv) & 65.5 & 65.0 & 66.1 \\ \bottomrule
\end{tabular}
\end{table}

\section{Details for Image Generation}
We train GMMNs on the benchmark datasets MNIST~\cite{lecun1998gradient}. We use the standard test set of 10,000 images, and randomly select 5000 from the standard 60,000 training images for validation. The remaining 55,000 are used for training. We train the GMMN network in both the input data space and code space of an auto-encoder. For all the networks, a uniform distribution in [-1, 1$]^{H}$ is used as the prior for the $H$-dimensional stochastic hidden layer at the top of the GMMN, which is followed by 4 ReLU layers. The output layer is a logistic sigmoid function, which guarantees that the code space dimensions lay in [0, 1]. The auto-encoder has 4 layers, 2 for the encoder and 2 for the decoder. For more details about the architecture of GMMN and auto-encoder, please refer to the original paper~\cite{li2015generative}. 

We train the GMMNs with mini-batch of size 1000, for each mini-batch, a set of 1000 samples will be generated from the network. The loss and gradient are computed from these 2000 samples. We replace the original square root loss function $\mathcal{L}_{\operatorname{MMD}}$ with $\mathcal{L}_{\operatorname{match}}$ of L2M to get the result GMMN+L2M. We set max epochs to be 500 and use Adam as the optimization strategy. The learning rate and momentum for both GMMN and auto-encoder, dropout rate for the auto-encoder are tuned using Bayesian optimization~\cite{snoek2012practical}.

\begin{figure}[ht!]
	\centering
		\subfigure[GMMN]{
			\centering
			\includegraphics[scale=0.3]{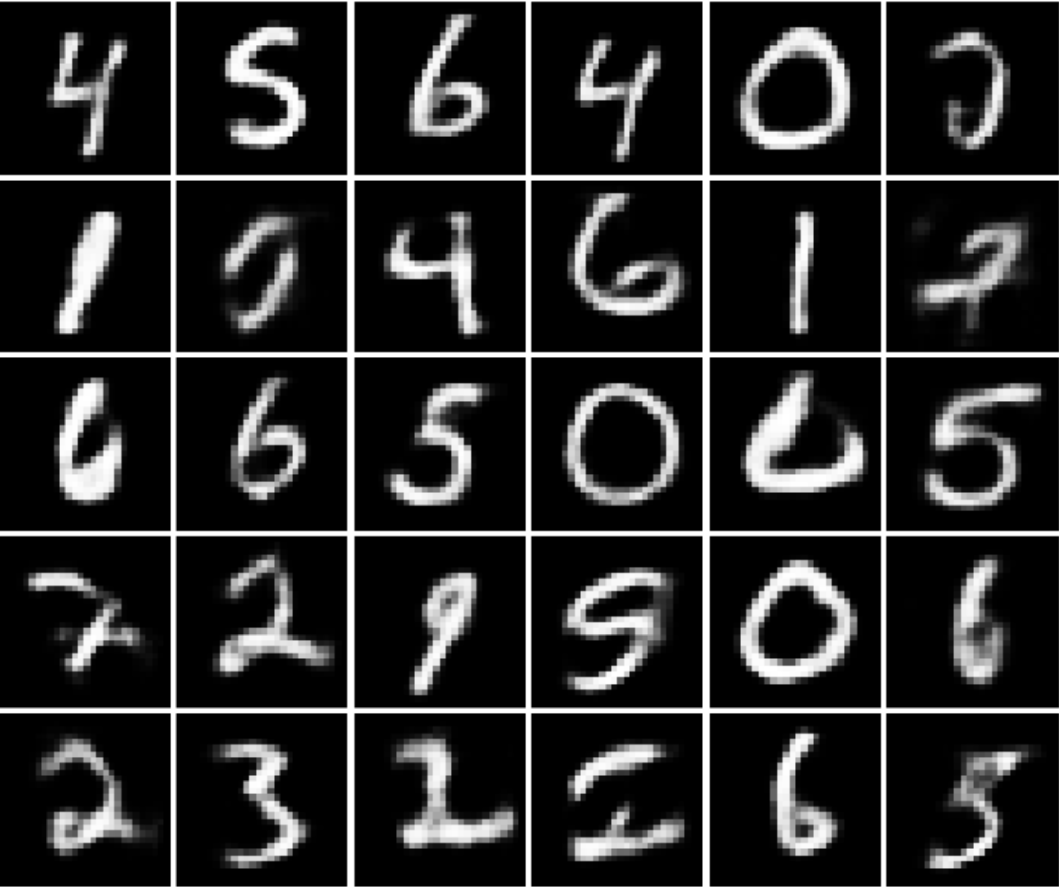}
			\label{fig-gmmn-mmd}}
		\subfigure[L2M]{
			\centering
			\includegraphics[scale=0.3]{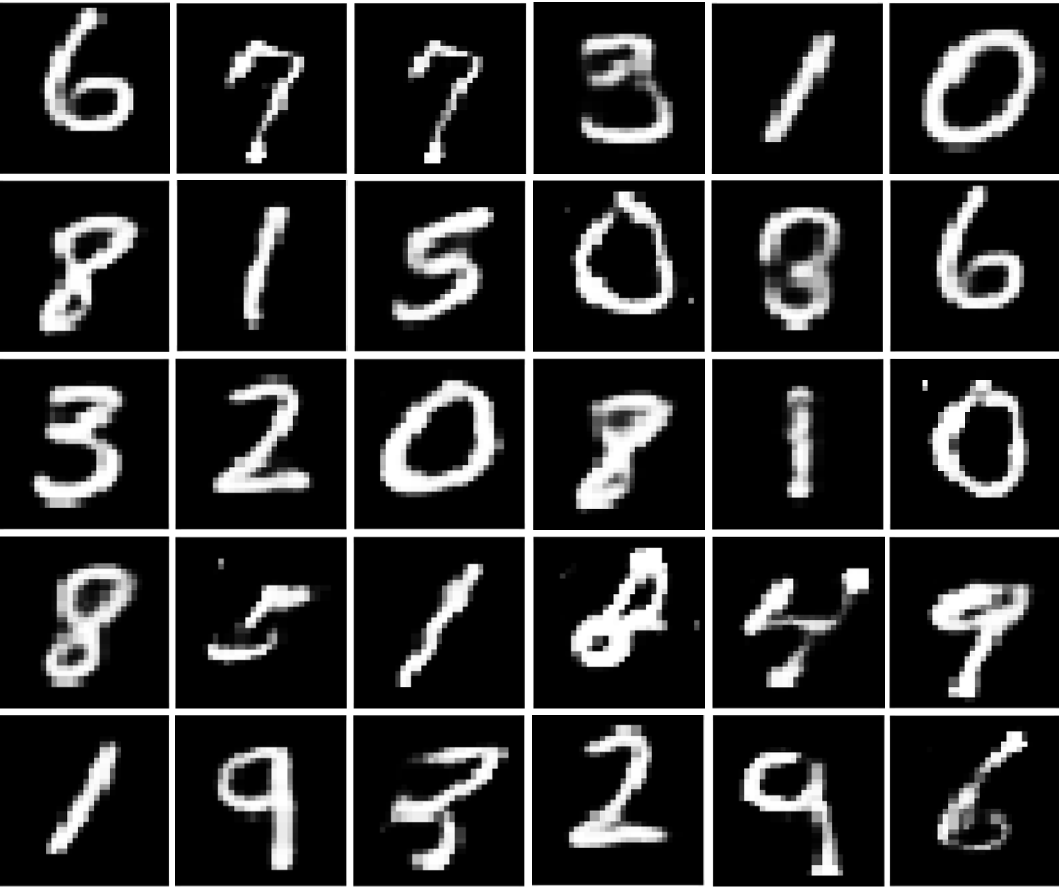}
			\label{fig-gmmn-l2m}}
	\caption{More MNIST samples generated by GMMN and L2M.}
	\label{fig-generative-more}
\end{figure}

Fig.~\ref{fig-generative-more} shows more MNIST samples generated by GMMN with MMD and L2M. It is clear that L2M generates sharper samples than MMD. We believe that L2M has more potential in image generation and this is only a test experiment. We are well aware that there are lots of existing works for image generation these years and adhere to hope that L2M could be significantly extended for this task in the future.

%\bibliographystyle{unsrt}
%\bibliography{nips20}

%\end{document}

\end{document}